\documentclass[runningheads]{llncs}
\pdfoutput=1
\usepackage{graphicx}
\usepackage{comment}
\usepackage{amsmath,amssymb} 
\usepackage{color}
\usepackage{booktabs}
\usepackage{multicol}
\usepackage{paralist} 
\usepackage{subcaption}
\usepackage{bm} 

\usepackage{algpseudocode,algorithm,algorithmicx}
\algnewcommand\algorithmicforeach{\textbf{for each}}
\algdef{S}[FOR]{ForEach}[1]{\algorithmicforeach\ #1\ \algorithmicdo}

\algrenewcommand\algorithmicensure{\textbf{Postcondition:}}

\newcommand{\StatexIndent}[1][3]{%
  \setlength\@tempdima{\algorithmicindent}%
  \Statex\hskip\dimexpr#1\@tempdima\relax}

\newcommand*\samethanks[1][\value{footnote}]{\footnotemark[#1]}
\usepackage[width=122mm,left=12mm,paperwidth=146mm,height=193mm,top=12mm,paperheight=217mm]{geometry}

\begin{document}
\pagestyle{headings}
\mainmatter
\def\ECCVSubNumber{}  

\title{OpenGAN: Open Set Generative Adversarial Networks} 

%
\author{Luke Ditria\thanks{Authors contributed equally.} \and
Benjamin J. Meyer\samethanks \and
Tom Drummond }
\authorrunning{L. Ditria et al.}
%
\institute{ARC Centre of Excellence for Robotic Vision, Monash University\\
\email{\{luke.ditria,benjamin.meyer,tom.drummond\}@monash.edu}}
\maketitle

\begin{abstract}

Many existing conditional Generative Adversarial Networks (cGANs) are limited to conditioning on pre-defined and fixed class-level semantic labels or attributes. We propose an open set GAN architecture (OpenGAN) that is conditioned per-input sample with a feature embedding drawn from a metric space. Using a state-of-the-art metric learning model that encodes both class-level and fine-grained semantic information, we are able to generate samples that are semantically similar to a given source image. The semantic information extracted by the metric learning model transfers to out-of-distribution novel classes, allowing the generative model to produce samples that are outside of the training distribution.
We show that our proposed method is able to generate 256$\times$256 resolution images from novel classes that are of similar visual quality to those from the training classes. In lieu of a source image, we demonstrate that random sampling of the metric space also results in high-quality samples. We show that interpolation in the feature space and latent space results in semantically and visually plausible transformations in the image space. Finally, the usefulness of the generated samples to the downstream task of data augmentation is demonstrated. We show that classifier performance can be significantly improved by augmenting the training data with OpenGAN samples on classes that are outside of the GAN training distribution.

\end{abstract}

\section{Introduction}

Generating new data that matches a target distribution is a challenging problem with applications including image-to-image translation \cite{isola2017image,zhu2017unpaired}, data augmentation \cite{frid2018gan,xing2019adversarial} and video prediction \cite{liang2017dual,kwon2019predicting}. A popular approach to this problem is Generative Adversarial Networks (GANs) \cite{goodfellow2014generative}, which train a generator and discriminator network in an adversarial manner. However, such networks have issues with training instability, especially for complicated and multi-modal data, and often result in a lack of diversity in the generated samples, particularly when training data is limited \cite{gurumurthy2017deligan}. 
Conditional GANs (cGANs) \cite{mirza2014conditional} achieve greater control over the generated samples by conditioning the model on information including class labels \cite{odena2017conditional,miyato2018cgans,zhang2018self,brock2019big}, attributes \cite{gauthier2014conditional,8100224,lu2018attribute}, textual attributes \cite{reed2016generative,reed2016learning,mansimov2015generating,stackgan,dong2017semantic,park2018mc} or object pose \cite{tran2017disentangled,ge2018fd,Zakharov_2019_ICCV}. However, class conditional GANs are unable to generate novel class samples, attribute conditional GANs are limited to a fixed set of pre-defined attributes and pose conditional GANs require hand-labelled and pre-defined pose codes or object landmarks. 
While some existing methods condition on image-level features using an encoder-decoder architecture \cite{antoniou2017data,tran2017disentangled,NIPS2018_7504}, these approaches train the encoder concurrently with the generator, enforcing no restrictions on the information encoded in the features. This can undesirably result in significant variation of the discriminative semantic information in samples generated from the same source image (see Section \ref{subsec:split}).

\begin{figure}[!t]
\centering
    \begin{subfigure}{0.8\linewidth}
        \begin{center}
        \begin{subfigure}{0.48\linewidth} 
            \includegraphics[width=\linewidth]{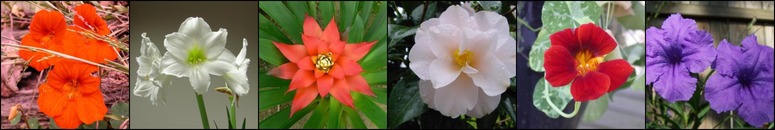}
        \end{subfigure}\hspace{8pt}%
        \begin{subfigure}{0.48\linewidth} 
            \includegraphics[width=\linewidth]{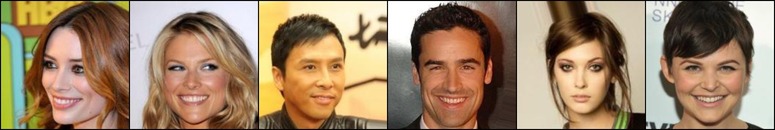}
        \end{subfigure}
        \begin{subfigure}{0.48\linewidth} 
            \includegraphics[width=\linewidth]{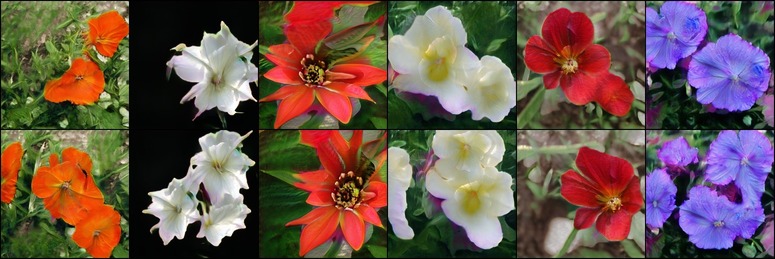}
        \end{subfigure}\hspace{8pt}%
        \begin{subfigure}{0.48\linewidth} 
            \includegraphics[width=\linewidth]{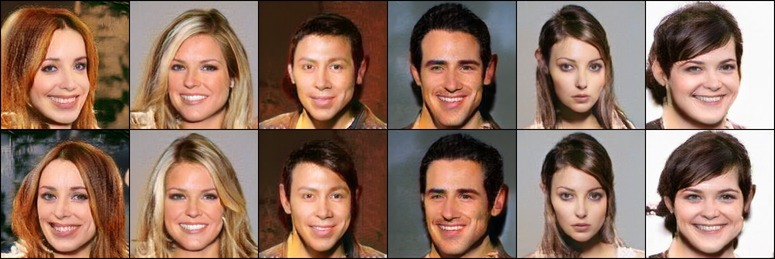}
        \end{subfigure}
        \end{center}
        \end{subfigure}
        \caption{Given novel class source images (top row of each section), our approach is able to generate 256$\times$256 samples (bottom two rows of each section) that closely match the features of the source. OpenGAN was not trained on the classes shown.}
        \label{fig:key_result}
\end{figure}

In this work, we propose an open set GAN (OpenGAN) that conditions the model on per-image features drawn from a metric space. Deep metric learning approaches have been shown to learn metric spaces that encode both class-level and fine-grained semantic information, and also have the ability to transfer to novel, out-of-distribution classes \cite{7780803,NIPS2016_6200,7298682,harwood2017smart,DBLP:journals/corr/SongJR016,rippel2016metric,meyer2018deep,7780803}.
By conditioning on per-image metric features, our proposed model is not limited to closed-set problems, but can also generate samples from novel classes in the open-set domain (see Figure \ref{fig:key_result}). Further, this conditioning method results in high intra-class diversity, where that is desirable.
Unlike many existing methods, our approach is not conditioned on class-level information alone or pre-defined attributes and poses, but rather on the semantic information extracted by a state-of-the-art deep metric learning model. Additionally, the proposed approach differs from existing feature matching GANs \cite{NIPS2016_6125,NIPS2006_3110,Gretton:2012:KTT:2188385.2188410}, as it does not attempt to match feature moments over the entire dataset, but conditions the model on a per-feature basis. During testing, data can be generated by conditioning on specific source images, or by randomly sampling the metric feature space.

Given a metric feature extracted from a real source image, our model generates images that visually and semantically match the source, as shown in Figure \ref{fig:key_result}. The generator should not simply reconstruct the source image, but produce images with features that are similar to the source, when passed through the metric learning model. 
Conditioning the generator on semantically rich features not only allows for the generation of both in-distribution and novel class images, but also for transfer between source domains (Section \ref{subsec:domain}). Further, OpenGAN samples can be successfully utilised for data augmentation (Section \ref{subsec:data_aug}).

The use of a metric learning model is an important design decision. Metric features describe only discriminative semantic information, ignoring all contextual and structural information, such as pose, the quantity and arrangement of objects and other non-discriminative intra-class and inter-class variations. As a result, the generator relies on a latent space noise vector to map this information (and only this information), meaning that the structural and contextual information can be modified without any variation occurring between the semantic content of the source image and generated image. This is unlike in encoder-decoder GAN architectures that learn to extract image features concurrently with the GAN \cite{antoniou2017data,tran2017disentangled,NIPS2018_7504}. For our approach, the content information is cleanly split into two distinct spaces, without the need for the pre-defined, hand-labelled pose and landmark information that is required in previous work \cite{tran2017disentangled,ge2018fd,Zakharov_2019_ICCV}.

\section{Related Work}

\subsubsection{Conditional Generative Models.}
The two most commonly used generative models in recent times are Generative Adversarial Networks \cite{goodfellow2014generative} and Variational Auto-Encoders \cite{kingma2013auto}. In this work, we focus on deep convolutional GANs \cite{radford2015unsupervised}. Several methods have been proposed to achieve greater control over the generated images. Mirza and Osindero \cite{mirza2014conditional} condition on class-level labels by supplying one-hot class vectors to both the generator and discriminator. Such an approach can improve both generated image quality and inter-class diversity in the generator distribution. Incorporating class-level information by treating the discriminator as a multi-label classifier has also been shown to improve the quality of generated samples \cite{odena2016semi,NIPS2016_6125}. Odena \textit{et al.} \cite{odena2017conditional} extend this by tasking the discriminator with estimating both the probability distribution over class labels and over the source distribution (i.e. real or fake).
Conditional information can also be incorporated by way of conditional normalisation layers \cite{dumoulin2016learned,de2017modulating,miyato2018cgans,zhang2018self,brock2019big}, which learn the batch normalisation \cite{pmlr-v37-ioffe15} or instance normalisation \cite{ulyanov2016instance} scale and bias terms as a function of some input.

Beyond class-level conditional information, data generation can also guided by conditioning the model on pre-defined attributes \cite{gauthier2014conditional,8100224,lu2018attribute}, such as hair colour and style for face generation. 
Similarly, generative models can be conditioned on attributes in a textual form by text-to-image synthesis methods \cite{reed2016generative,reed2016learning,mansimov2015generating,stackgan,dong2017semantic,park2018mc}. GANs can be conditioned on structural information, allowing direct control over the object pose in the generated image. Such methods require hand-labelled and pre-defined pose codes or object landmarks \cite{tran2017disentangled,ge2018fd,Zakharov_2019_ICCV}.

Methods including DAGAN \cite{antoniou2017data}, MetaGAN \cite{NIPS2018_7504} and DR-GAN \cite{tran2017disentangled} use an encoder-decoder structure, allowing the generator to be conditioned on image-level features. As such, these approaches are not limited to in-distribution classes by their design, unlike class conditional GANs. However, the encoder and generator are trained simultaneously with no constraints on the information that is represented in the encoder features.
Unlike these methods, our approach leverages metric features extracted from a deep metric learning model that is trained prior to the GAN.
Metric learning models have a demonstrated efficacy for open set problems \cite{8794188}, as such, our approach is explicitly designed for the open set domain. Further unlike the encoder-decoder GANs, our method results in no semantic variation when changing only the latent vector. This is because all discriminative semantic information is encoded in the metric features and the generator is constrained to produce images with features that match those of the source image. Consequently, the latent vector can only encode non-discriminative information, such as the object pose, the background and the number of objects in the image. Encoder-decoder GANs do not enforce these feature constraints.

Nguyen \textit{et al.} \cite{Nguyen:2016:SPI:3157382.3157477,nguyen2017plug} condition the generator using an auxiliary classifier network by finding the latent vector that results in generated data that strongly activates neurons in the auxiliary network. These so called Plug and Play Generative Networks can generate data that is outside of the generator's training distribution, but is inside the auxiliary network's training distribution. For our approach, the generator and feature extractor training distributions are the same, and the generator can be conditioned on data that is outside of that distribution.

\subsubsection{Matching Networks.} Training stability of GANs can be improved by performing feature matching \cite{NIPS2016_6125}. The generator is trained such that the expected value of features extracted from generated data by a given layer of the discriminator matches that of the real data. Similar to feature matching networks are moment matching networks \cite{li2015generative,Dziugaite:2015:TGN:3020847.3020875,li2017mmd}, which generally try to match all moments of the distributions using maximum mean discrepancy \cite{NIPS2006_3110,Gretton:2012:KTT:2188385.2188410}. Unlike feature matching networks, our approach attempts to match per-sample source features individually, rather than the expected value. Our generator is also directly conditioned on per-sample features, such that the generated samples match the semantic content of the source features. Further, we use an auxiliary metric learning model to extract features, rather than the discriminator. 
Our feature matching is also related to perceptual loss functions \cite{Johnson2016PerceptualLF}, which use a pre-trained classifier network to match the low and high level features of input and target images for problems such as style transfer. 

\subsubsection{Metric Learning.}
Rich visual features can be extracted from images by using a deep convolutional neural network to learn a distance metric over the images \cite{7780803,NIPS2016_6200,7298682,harwood2017smart,DBLP:journals/corr/SongJR016,rippel2016metric,meyer2018deep,7780803}. Many of these so-called deep metric learning methods are based on Siamese \cite{14fd455cbe634a189e2ed78a1b3a918c,1467314,1640964} and triplet networks \cite{DBLP:journals/corr/HofferA14}, which perform distance comparisons in the feature space. Research in this area often focuses on the generalisation of triplet loss, such that multiple pairwise distance comparisons can be made for a given example within a training batch \cite{7780803,NIPS2016_6200}. Other work focuses on the selection of informative triplets via mining techniques \cite{7298682,harwood2017smart}. Beyond triplet loss, the work by Song \textit{et al.} \cite{DBLP:journals/corr/SongJR016} directly minimises a clustering measure. Rippel \textit{et al.} propose Magnet loss \cite{rippel2016metric}, which explicitly models class distributions in the feature space and penalises class overlap. Other approaches minimise Neighbourhood Component Analysis (NCA) loss over the set of training features \cite{meyer2018deep} or per-class proxy features \cite{Movshovitz2017proxies}.
Metric learning has been combined with generative models to improve the stability of GAN training \cite{dou2017metric,dai2017metric}, as well as to improve the training of a metric learning model \cite{zieba2017training}. Unlike these methods, we use a metric learning model to condition a GAN on image features.

\section{Background}
\subsection{Generative Adversarial Networks}

Let $G$ be a \textit{generator} network that attempts to learn a mapping from a latent space to a target data space. Specifically, an image is generated as $\bar{\mathbf{x}} = G(\mathbf{z})$, where $\mathbf{z}$ is a latent vector sampled from the distribution $p_z = \mathcal{N}(0,1)$. Further, let $D$ be a \textit{discriminator} network that takes as input an image and attempts to distinguish between the generator distribution and the real data distribution $p_d$. The two networks are trained in an adversarial fashion, with improvement in one network driving improvement in the other. Greater control over the generated image can be achieved by conditioning both networks on a label $y \in p_d$.

\subsection{Deep Metric Learning} \label{sec:dml}

\begin{figure}[t]
\begin{center}
\includegraphics[width=0.3\linewidth]{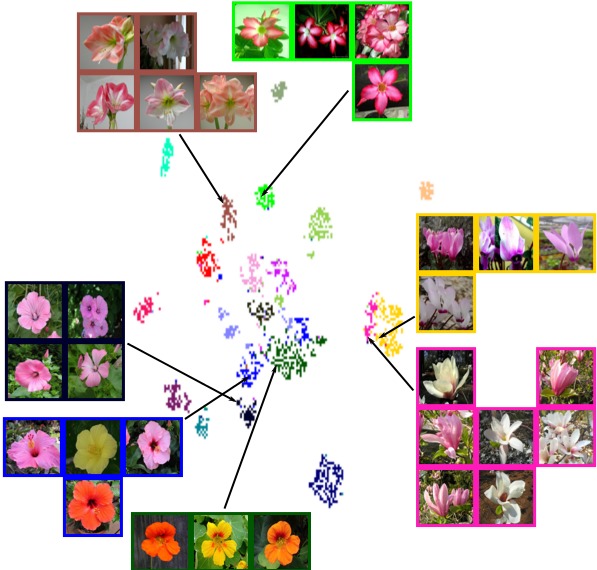}
\end{center}
   \caption{Visualisation of the metric feature space for 20 novel classes, represented by colour. The feature extractor is not trained on any of the shown flower species, yet examples are co-located based on class. Example images are selected to show similar classes being located nearby. Best viewed zoomed-in.}
\label{fig:real_popout}
\end{figure}

Our proposed method is agnostic in terms of the metric learning model used to extract image features. Here, we detail the metric learning algorithm that is used for all experiments in Section \ref{sec:experiments}. This particular approach \cite{meyer2018deep} is selected due to its ability to extract semantically rich features that both transfer well to novel classes and encode fine-grained intra-class and inter-class variations. This is shown in the t-SNE visualisation \cite{maaten2008visualizing} of novel class examples in Figure \ref{fig:real_popout}. Despite being from outside of the training set distribution, examples are well clustered based on class, with semantically similar classes located nearby.

For a given input image, the network $F$ extracts a $d$-dimensional feature $\mathbf{f} = [f^{(i)},...,f^{(d)}]$.
For training, a set of $n$ Gaussian kernel centres are defined in the feature space as $\mathcal{C} = \{\mathbf{c}_1,...,\mathbf{c}_n\}$, where $\mathbf{c}_i = [c_i^{(1)},...,c_i^{(d)}]$ is the $i$-th kernel centre. The centres are defined to be the locations of the $n$ training set features, with the weights of $F$ updated during training by minimising the NCA loss \cite{goldberger2005neighbourhood}. To make training feasible, a cached version of the kernel centres $\hat{\mathcal{C}}$ is stored and updated periodically during training, avoiding the need to do so at every training iteration. The loss minimised during training is shown in Equation \ref{eq:nca_loss},  where $\sigma$ is a hyperparameter and $\ell_i$ is the class label of the $i$-th training example. If necessary, approximate nearest neighbour search can leveraged to make the approach scalable both in terms of the number of classes and training examples.
\begin{equation} \label{eq:nca_loss}
loss_F = - \sum_{\mathbf{c}_i \in \mathcal{C}}\ln \left( \frac{ \sum_{\hat{\mathbf{c}}_j \in \hat{\mathcal{C}}, i \neq j, \ell_i = \ell_j} \exp{\left(\frac{-\lVert \mathbf{c}_i - \hat{\mathbf{c}}_j \rVert^2}{2\sigma^2} \right)}} { \sum_{\hat{\mathbf{c}}_k \in \hat{\mathcal{C}}, i \neq k} \exp{\left(\frac{-\lVert \mathbf{c}_i - \hat{\mathbf{c}}_k \rVert^2}{2\sigma^2} \right)} }\right)
\end{equation}

\begin{figure}[t]
\centering
\begin{subfigure}{0.4\linewidth}
    \centering
    \includegraphics[width=0.8\linewidth]{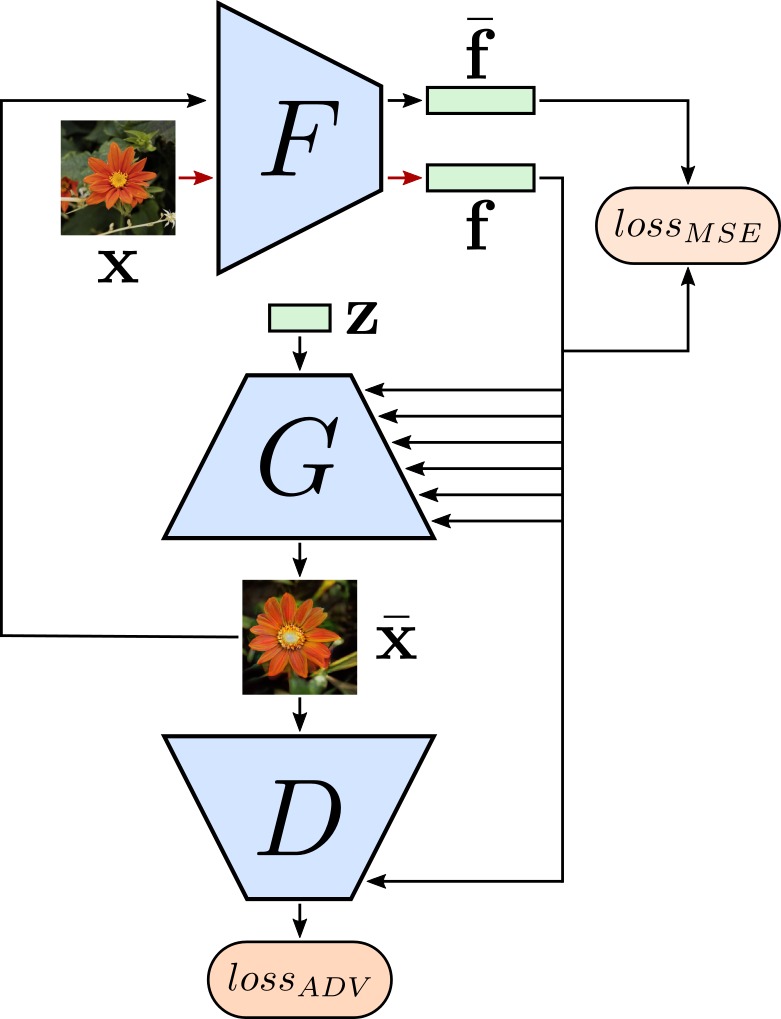}
    \caption{Overview of method.}
    \label{fig:overview}
\end{subfigure}\hspace{12pt}%
\begin{subfigure}{0.5\linewidth}
    \centering
        \begin{subfigure}{0.76\linewidth} 
            \centering
            \includegraphics[width=0.74\linewidth]{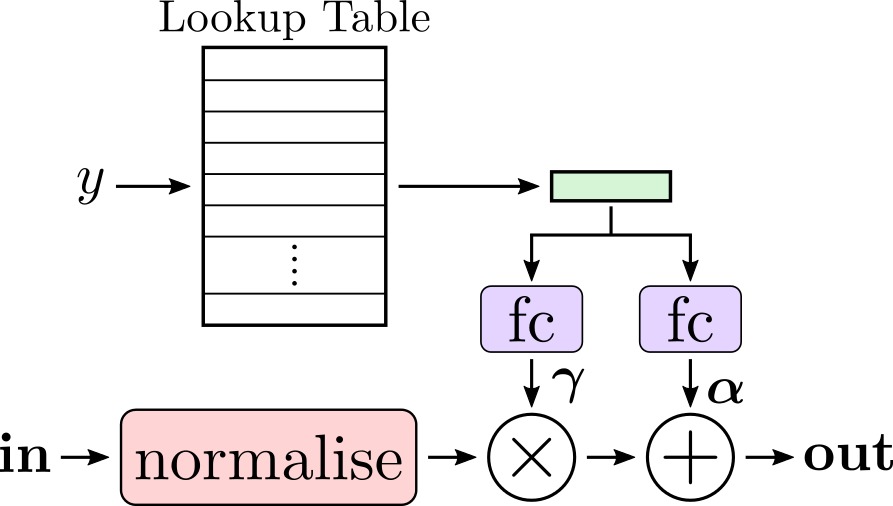}
            \caption{Class conditional normalisation.}
            \label{subfig:cc}
        \end{subfigure}\vspace{4pt}
        \begin{subfigure}{0.76\linewidth} 
            \centering
            \includegraphics[width=0.74\linewidth]{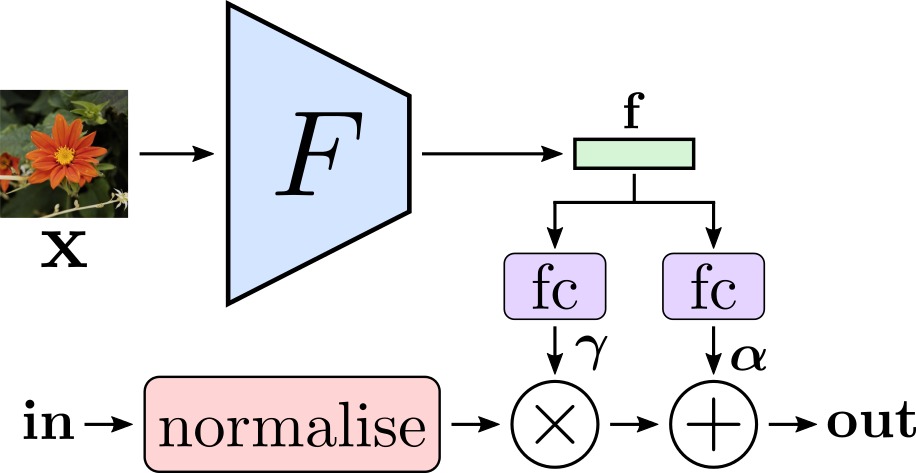}
            \caption{Feature embedding conditional normalisation.}
            \label{subfig:fc}
    \end{subfigure}
\end{subfigure}

   \caption{(a) Overview of our approach. During training, features are extracted from training images and used to condition the GAN via conditional normalisation layers. During testing, features may be extracted from source images or randomly sampled from the metric space. 
   (b) Class conditional normalisation compared to the feature conditional normalisation (c) used in our model.}

\end{figure}

\section{Feature Embedding Conditional GANs}

\subsection{Overview} \label{subsec:overview}
Our proposed method consists of three convolutional neural networks: a generator $G$, a discriminator $D$ and a metric feature extractor $F$. Network $F$ extracts a feature $\mathbf{f}$ for a sampled training image, which is fed into networks $G$ and $D$. The generator attempts to produce an image that both fools the discriminator and results in a feature $\bar{\mathbf{f}}$ that closely matches the real feature, when the fake image is passed through network $F$. The former is achieved via an adversarial (ADV) loss, while the latter is achieved by a mean squared error (MSE) loss term in the metric feature space. This is illustrated in Figure \ref{fig:overview}. During testing, examples can be generated by conditioning on features from specific images or by simply sampling the feature space.

Features are incorporated into the generator and discriminator by way of feature conditional normalisation layers, described in detail in Section \ref{subsec:fcbn}. The normalisation scale and bias terms are learned as a continuous function of the conditioning features, as opposed to a discreet function of class labels in class conditional normalisation. This continuity means that during testing, meaningful interpolation between features can occur. Further, out-of-distribution images can be generated by sampling a desired point in the metric feature space or by conditioning on the feature extracted from a specific novel image.

In a conventional GAN or class conditional GAN framework, generating an image that visually and semantically matches a given source image can be challenging. 
Additionally, there is no mechanism in the generator training that encourages the ability to transfer to data outside of the training distribution. Conversely, the training of our generator is guided by a feature extractor that transfers to novel classes (see Section \ref{sec:dml}) and the ability to condition the generator on a specific source image is built-in to the framework.

\begin{algorithm}[t]
  \caption{Training algorithm for OpenGAN.
    \label{algo}}
    \begin{multicols}{2}
  \begin{algorithmic}[1]
	\Require
		\Statex Models F, G, D with parameters $\bm{\theta}_F$, $\bm{\theta}_G$, $\bm{\theta}_D$
		\Statex Scale term for feature loss $\lambda$
	\State Pre-train $\bm{\theta}_F$ (Section \ref{sec:dml})
	\While {$\bm{\theta}_G$ is not converged}
	    \State Sample $\mathbf{x} \sim p_d $
	    \State $\mathbf{f} \gets F(\mathbf{x})$
	    \State Sample $\mathbf{z} \sim \mathcal{N}(0,1)$
	    \vfill\null
        \columnbreak
	    \State $\mathcal{L}_D \gets \min (0, 1-D(\mathbf{x},\mathbf{f})) $
	    \Statex $+ \min (0, 1+D(G(\mathbf{z}, \mathbf{f}),\mathbf{f}))$
	    \State $\bm{\theta}_D \gets \bm{\theta}_D - \mathrm{Adam}(\nabla \mathcal{L}_D)$ 
	    \State Sample $\mathbf{z} \sim \mathcal{N}(0,1)$
	    \State $\bar{\mathbf{x}} \gets G(\mathbf{z}, \mathbf{f})$
	    \State $\mathcal{L}_G \gets -D(\bar{\mathbf{x}}, \mathbf{f}) + \lambda \: \left\lVert F(\bar{\mathbf{x}}) - \mathbf{f} \right\rVert^2$
	    \State $\bm{\theta}_G \gets \bm{\theta}_G - \mathrm{Adam}(\nabla \mathcal{L}_G)$
	\EndWhile
	\Statex
\end{algorithmic}
\end{multicols}
\end{algorithm}

\subsection{Training Procedure} \label{subsec:training}

Network optimisation is outlined in Algorithm \ref{algo}. The feature extractor is pre-trained, with the weights subsequently frozen. The loss functions minimised by the discriminator and the generator, respectively, are:
\begin{align}
    loss_D = \: & \mathbb{E}_{\mathbf{x} \sim p_d} \left[ \min(0,1 - D(\mathbf{x}, \mathbf{f})) \right] + 
             \mathbb{E}_{\mathbf{x} \sim p_d, \mathbf{z} \sim p_z} \left[ \min(0,1+D(G(\mathbf{z}, \mathbf{f}), \mathbf{f})) \right], \label{eq:fc_hinge_d} \\
    loss_G = \: & \mathbb{E}_{\mathbf{x} \sim p_d, \mathbf{z} \sim p_z} \big[ -D(G(\mathbf{z}, \mathbf{f}), \mathbf{f}) + \lambda \: \left\lVert F(G(\mathbf{z}, \mathbf{f})) - \mathbf{f} \right\rVert^2 \big],  \label{eq:fc_hinge_g}
\end{align}
where $\lambda$ is a scaling term for the feature loss component and $\mathbf{f} = F(\mathbf{x})$ . Hinge loss \cite{lim2017geometric,tran2017hierarchical} is used for the adversarial component of the losses, while mean squared error is used for the feature loss component. Model parameters are updated by gradient descent with Adam optimisation \cite{kingma2014adam}.

\subsection{Feature Conditional Normalisation} \label{subsec:fcbn}

Intermediate neural network layer activations can be forced to have similar distributions by including layers that normalise over the entire batch \cite{pmlr-v37-ioffe15} or over each instance individually \cite{ulyanov2016instance}. Normalisation of activations can lead to faster and more stable training, as well as better overall model performance. Such layers perform the following normalisation on an activation:
\begin{equation}
    \hat{m}_i = \gamma\frac{m_i - \mu}{\sqrt{v+\epsilon}} + \alpha
\end{equation}
where $m_i$ is the input, $\hat{m}_i$ is the normalised output, $\mu$ is the mean, $v$ is the variance and $\epsilon$ is a small constant. In conventional normalisation layers, the scale $\gamma$ and bias $\alpha$ terms are learned model parameters, while for conditional normalisation layers, they are learned as a function of some input. Class conditional normalisation (Figure \ref{subfig:cc}) learns a feature per-class that is often input to two fully connected (FC) layers to produce the scale and bias terms. This limits the network to produce only images from the training distribution or to an interpolation between training classes.

We propose metric feature embedding conditional normalisation (Figure \ref{subfig:fc}), which learns the scale and bias terms as a function of a feature embedding drawn from a metric space. This allows conditioning on specific images or features, compared to in-distribution class-level conditioning. 

\begin{table}[!t]
\centering
\caption{Comparison of FID and intra-class FID scores. Lower scores indicate better sample quality.}
\label{table:fid}
\begin{tabular}{lcc}
\toprule 
  & FID & Intra FID \\
\midrule
U-SAGAN \cite{zhang2018self} & 161.74 & -  \\
C-SAGAN \cite{zhang2018self} & 66.12 & 179.67 \\
Ours: T-SM & 22.05 & 103.18  \\
Ours: N-SM & 39.51 & 110.04  \\
Ours: N-RF & 31.89 & 104.90 \\
\bottomrule
\end{tabular}
\end{table}

\begin{figure*}[!t]
    \begin{center}
    \begin{subfigure}[t]{0.19\linewidth} 
        \includegraphics[width=\linewidth]{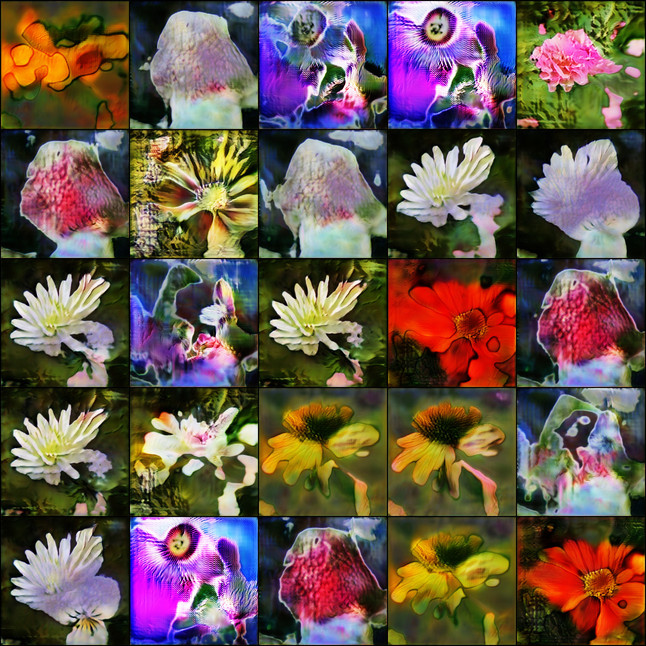}
        \caption{U-SAGAN.}
    \end{subfigure}\hspace{4pt}%
    \begin{subfigure}[t]{0.19\linewidth} 
        \includegraphics[width=\linewidth]{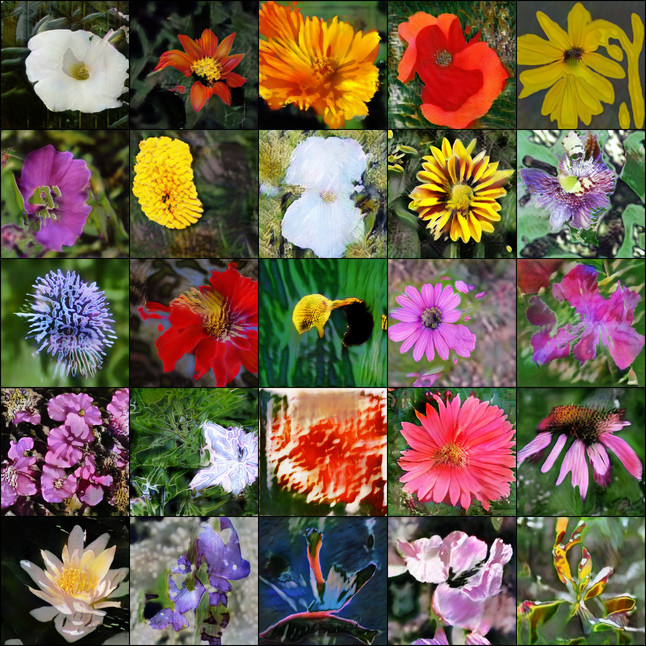}
        \caption{C-SAGAN.}
    \end{subfigure}\hspace{4pt}%
    \begin{subfigure}[t]{0.19\linewidth} 
        \includegraphics[width=\linewidth]{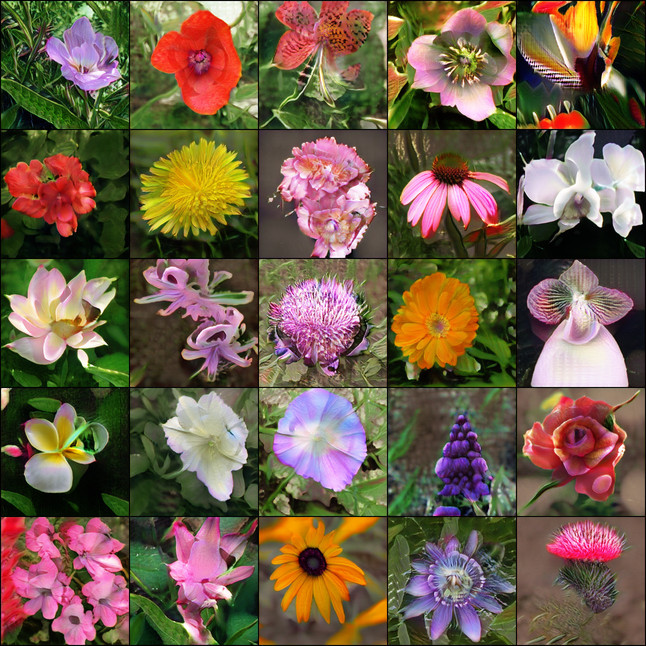}
        \caption{Ours: T-SM.}
    \end{subfigure}\hspace{4pt}%
    \begin{subfigure}[t]{0.19\linewidth} 
        \includegraphics[width=\linewidth]{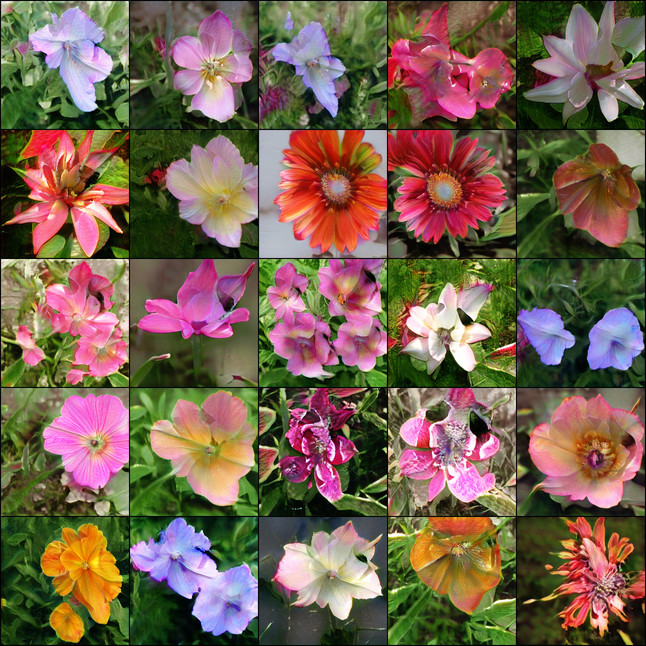}
        \caption{Ours: N-SM.}
    \end{subfigure}\hspace{4pt}%
    \begin{subfigure}[t]{0.19\linewidth} 
        \includegraphics[width=\linewidth]{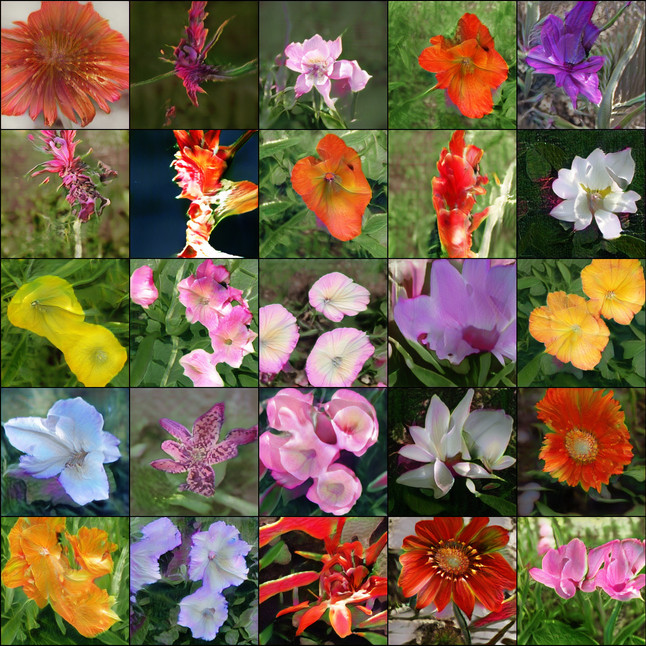}
        \caption{Ours: N-RF.}
    \end{subfigure}
    \end{center}
    \caption{Uncurated and randomly selected images on the Flowers102 dataset.}
    \label{fig:uncurated}
\end{figure*}

\begin{figure}[!t]
    \begin{center}
    \begin{subfigure}{0.6\linewidth} 
        \centering
        \begin{subfigure}[t]{0.32\linewidth} 
            \includegraphics[width=\linewidth]{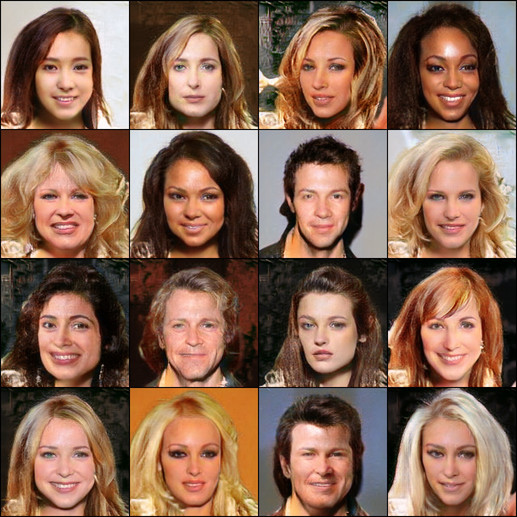}
            \caption{Ours: T-SM.}
            \label{subfig:uncurated_known_means}
        \end{subfigure}\hspace{4pt}%
        \begin{subfigure}[t]{0.32\linewidth} 
            \includegraphics[width=\linewidth]{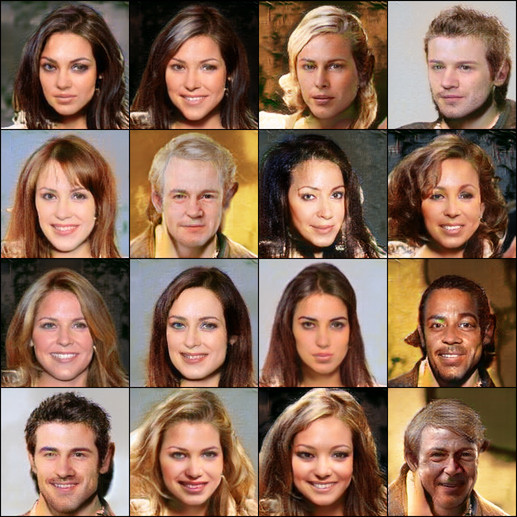}
            \caption{Ours: N-SM.}
            \label{subfig:uncurated_novel_means}
        \end{subfigure}\hspace{4pt}%
        \begin{subfigure}[t]{0.32\linewidth} 
            \includegraphics[width=\linewidth]{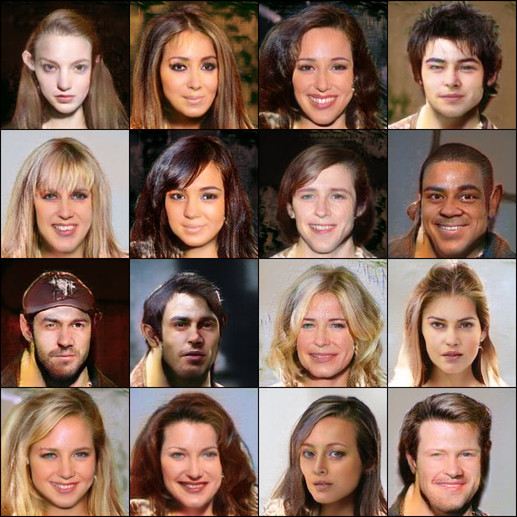}
            \caption{Ours: N-RF.}
            \label{subfig:uncurated_novel_ims}
        \end{subfigure}
    \end{subfigure}
    \end{center}
    \caption{Uncurated and randomly selected images on the CelebA dataset.}
    \label{fig:uncurated_faces}
\end{figure}

\begin{figure*}[!t]
    \begin{center}
    \begin{subfigure}{\linewidth} 
        \includegraphics[width=\linewidth]{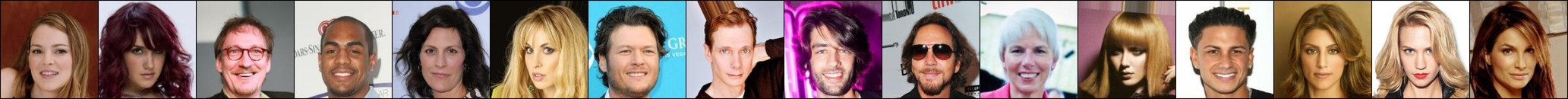}
    \end{subfigure}\vspace{1pt}
    \begin{subfigure}{\linewidth} 
        \includegraphics[width=\linewidth]{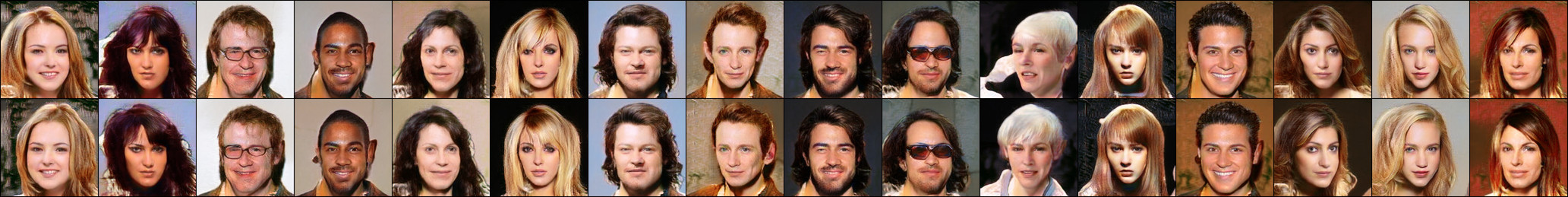}
    \end{subfigure}
    \end{center}
    \caption{Novel class real source images (top row) and resultant generated images (bottom two rows). Although the identities are not present during training, the fake images match the features of the real source images.}
    \label{fig:faces_vis_dist}
\end{figure*}

\section{Experiments} \label{sec:experiments}

\subsection{Implementation Details}

The datasets used for evaluation are Oxford Flowers102 \cite{Nilsback08} and CelebA Faces \cite{liu2015faceattributes}. Each dataset is split into training and novel classes. The first 82 classes from Flowers102 are used for training, resulting in 6433 training images. For CelebA, identities are used as class labels with the 3300 identities containing the most samples used for training, resulting in 97262 training images. The attribute and pose labels of CelebA are used for attribute and pose interpolation, however, the networks are not trained on this information.

The generator and discriminator follow a similar architecture to Self-Attention GAN (SAGAN) \cite{zhang2018self}, but we replace the projection layer in the discriminator with a single feature embedding conditional normalisation layer and a fully connected layer. We also generate images twice the resolution of SAGAN at 256$\times$256 pixels and use a channel width multiplier of 32. Six residual blocks \cite{7780459} are used in each network, along with spectral normalisation \cite{miyato2018spectral} and a single self-attention block \cite{zhang2018self}. Feature conditional normalisation is used in all residual blocks in the generator but only in the final discriminator block. We find batch normalisation on Flowers102 and instance normalisation on CelebA performs best in practice. A latent space dimension of 128 is used for the generator. For training, a base learning rate of $10^{-4}$, batch size of 48 and Adam optimiser \cite{kingma2014adam} with $\beta_1 = 0$ and $\beta_2 = 0.999$ are used. The value of $\lambda$ is set to 0.01. The GAN is trained for up to 60000 iterations on four Nvidia 1080 Ti GPUs, taking approximately 15 hours.

\begin{figure}[!t]
    \begin{center}
    \begin{subfigure}[b]{0.48\linewidth}
        \centering
        \begin{subfigure}{0.8\linewidth}
            \begin{subfigure}[b]{0.16\linewidth} 
                \includegraphics[width=\linewidth]{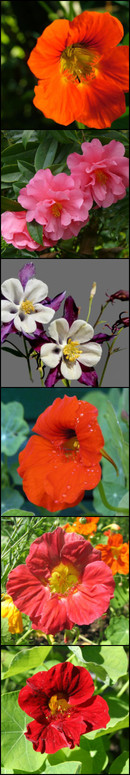}
            \end{subfigure}\hspace{0.04\linewidth}%
            \begin{subfigure}[b]{0.8\linewidth} 
                \includegraphics[width=\linewidth]{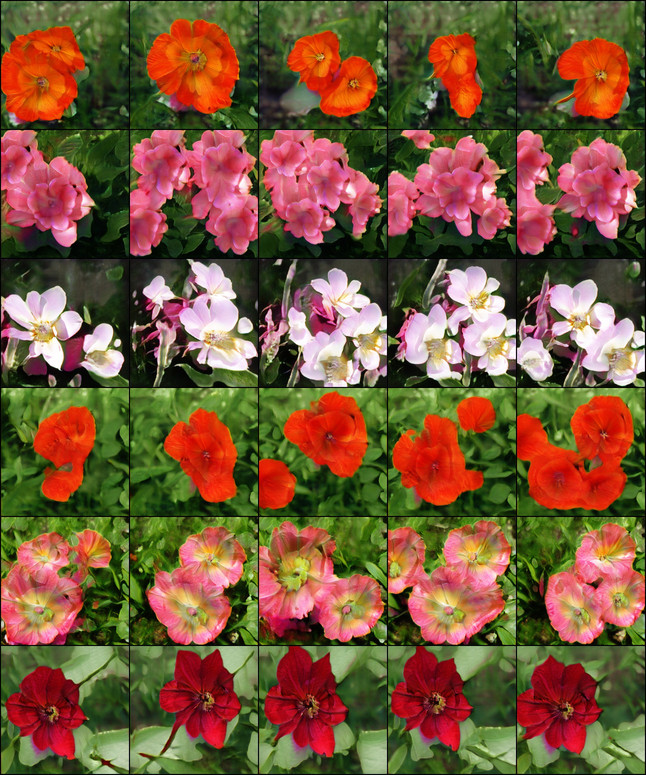}
            \end{subfigure}
        \end{subfigure}
        \caption{}
        \label{fig:vis_dist}
    \end{subfigure}%
    \begin{subfigure}[b]{0.48\linewidth} 
    \centering
        \begin{subfigure}[b]{0.75\linewidth} 
            \centering
            \includegraphics[width=\linewidth]{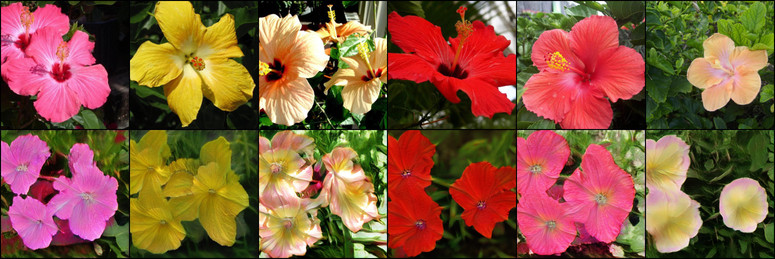}
        \end{subfigure}\vspace{2pt}
        \begin{subfigure}[b]{0.75\linewidth} 
            \centering
            \includegraphics[width=\linewidth]{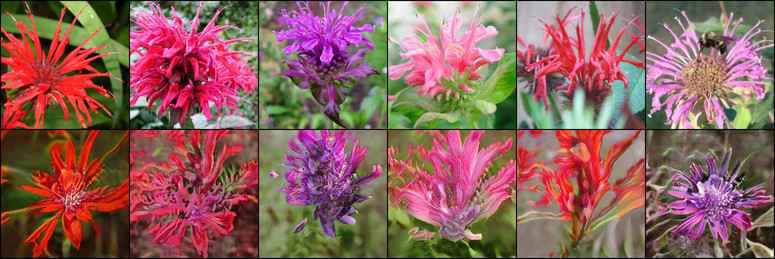}
        \end{subfigure}\vspace{2pt}
        \begin{subfigure}[b]{0.75\linewidth} 
            \centering
            \includegraphics[width=\linewidth]{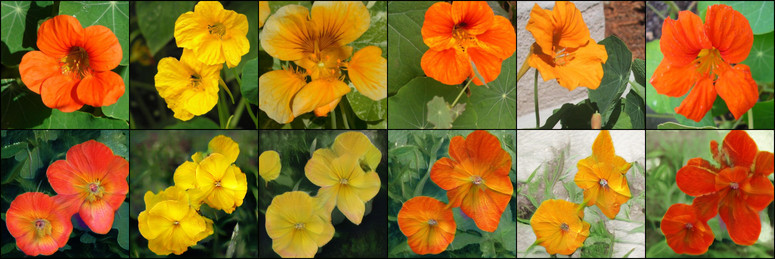}
        \end{subfigure}
        \caption{}
        \label{fig:fixed_z}
    \end{subfigure}
    \end{center}
    \caption{(a) Fake samples (right) are generated from a fixed feature, extracted from the novel class real samples (left).
    (b) Examples from three novel classes with a fixed latent vector for each class. The features of the real images (top rows of each section) are used to condition the GAN to produce the fake images (bottom rows of each section).}
    \label{fig:latentvsfeature}
\end{figure}

A ResNet18 architecture \cite{7780459} with the class-dependent fully connected layer removed is used for the feature extractor, producing 512-dimensional features. The model is trained with a base learning rate of $10^{-5}$, Gaussian $\sigma$ of 10 and an Adam optimiser \cite{kingma2014adam} with $\beta_1 = 0.9$ and $\beta_2 = 0.999$. The stored Gaussian centres are updated every 5 epochs.

\subsection{Comparison to Baselines} \label{subsec:comparison}

As the proposed method can use any suitable network architecture, the aim of this work is not to improve on state-of-the-art methods in terms of sample quality. Here, we aim to show that our method results in samples of at least comparable quality to appropriate baselines. We compare to two baselines: Unconditional SAGAN (\textit{U-SAGAN}) and Class Conditional SAGAN (\textit{C-SAGAN}) \cite{zhang2018self}. For fair comparison, these baselines have the same structure as our model, differing only in terms of the normalisation layers. U-SAGAN uses non-conditional normalisation, while C-SAGAN uses a single conditioning feature per-class (Figure \ref{subfig:cc}). 

\begin{figure}[!t]
    \begin{center}
    
    \begin{minipage}{0.44\linewidth} 
    \centering
        \includegraphics[width=\linewidth]{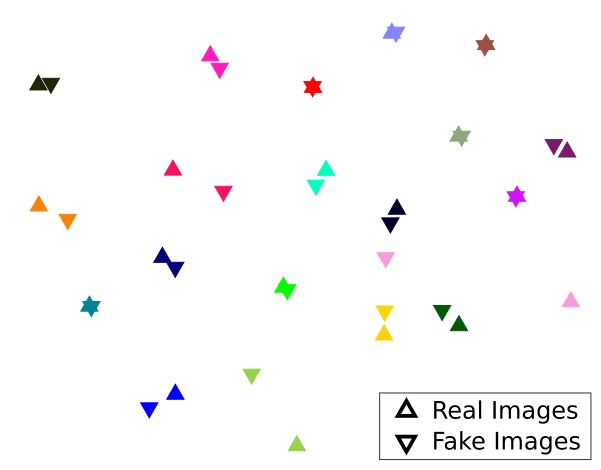}
       \caption{Novel per-class mean feature embeddings for real and fake images. Colour represents class.}
        \label{fig:fig_flowers_means}
    \end{minipage}\hspace{30pt}%
    \begin{minipage}{0.44\linewidth} 
    \centering
            \includegraphics[width=0.8\linewidth]{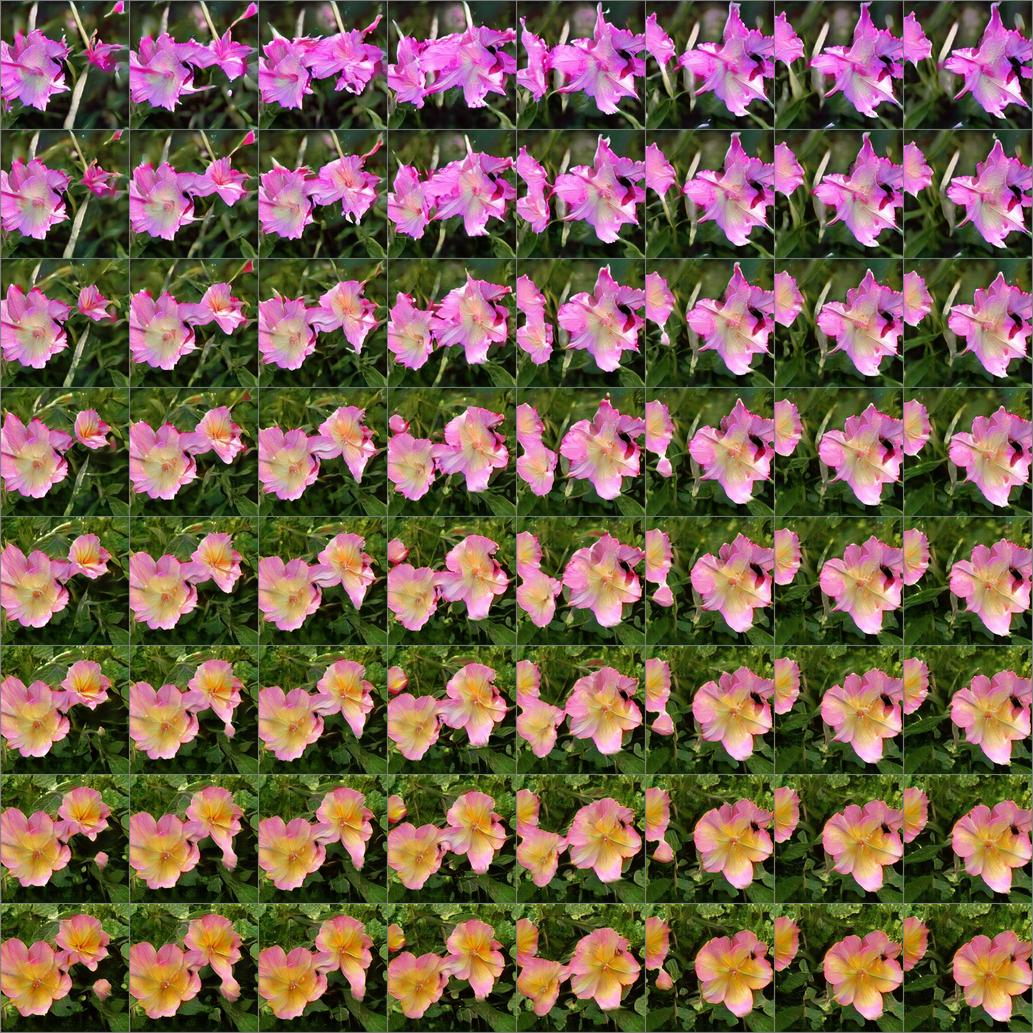}
        \caption{Interpolation between two latent vectors (horizontal) and two feature embeddings (vertical).}
        \label{fig:flowers_2d}
    \end{minipage}
    \end{center}
    
\end{figure}

Uncurated qualitative results on the Flowers102 dataset can be seen in Figure \ref{fig:uncurated} and a quantitative comparison, in terms of the FID and intra-class FID scores \cite{Heusel:fid}, is shown in Table \ref{table:fid}.
For our approach, we investigate sampling features from both the training and novel distributions, as well as two methods of feature sampling: random sampling from normal distributions the centred on the class means, and extracting features from sampled real images. Our methods are:
\begin{compactitem}
    \item \textit{Ours: T-SM}: Training distribution, sample means.
    \item \textit{Ours: N-SM}: Novel distribution, sample means.
    \item \textit{Ours: N-RF}: Novel distribution, real image features.
\end{compactitem}
Both qualitatively and quantitatively, there is little difference in quality between sampling training and novel distributions, or between the two feature sampling methods. This is also observed on CelebA in Figure \ref{fig:uncurated_faces}. Compared to the baselines, our approach results in both higher quality images and better sample diversity.

\subsection{One-Shot Image Generation} \label{subsec:one_shot}

In this section, we show that our method is able to generate samples that match the semantic features of source images sampled from the novel distribution. We name this problem ``one-shot image generation", however, it is important to note that no updates are made to the network weights using the novel source images; the source images are simply used to condition the generator. Figure \ref{fig:key_result} demonstrates this ability on both datasets, while further CelebA samples are shown in Figure \ref{fig:faces_vis_dist}. Additional Flowers102 samples are shown in Figures \ref{fig:vis_dist} and \ref{fig:fixed_z}, with discussion in Section \ref{subsec:diversity}.

Figure \ref{fig:fig_flowers_means} shows a t-SNE visualisation \cite{maaten2008visualizing} of the novel per-class mean features of the real and fake samples when passed through network $F$. In the majority of cases, the fake mean feature is co-located with the real mean feature.

\begin{figure}[!t]
    \begin{center}
    \begin{subfigure}{0.48\linewidth}
        \begin{subfigure}{\linewidth} 
            \centering
            \includegraphics[width=\linewidth]{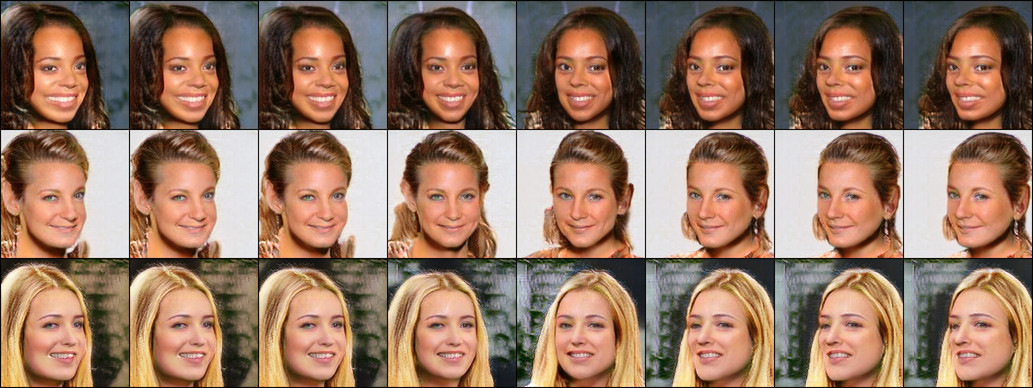}
            \caption{Pose interpolation in latent space. \phantom{adgdgdgdgd enej eknr}}
        \end{subfigure}\vspace{8pt}
        \begin{subfigure}{\linewidth} 
            \centering
            \includegraphics[width=\linewidth]{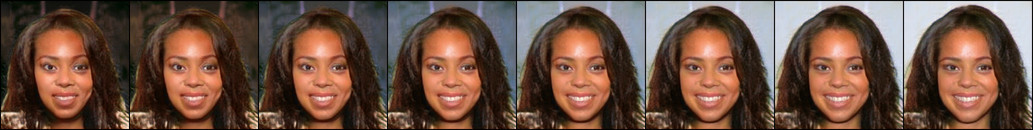}
            \caption{Pose interpolation in feature space. \phantom{adgdgdgdgd enej eknr}}
        \end{subfigure}
    \end{subfigure}\hspace{4pt}%
    \begin{subfigure}{0.48\linewidth}
        \begin{subfigure}{\linewidth} 
            \centering
            \includegraphics[width=\linewidth]{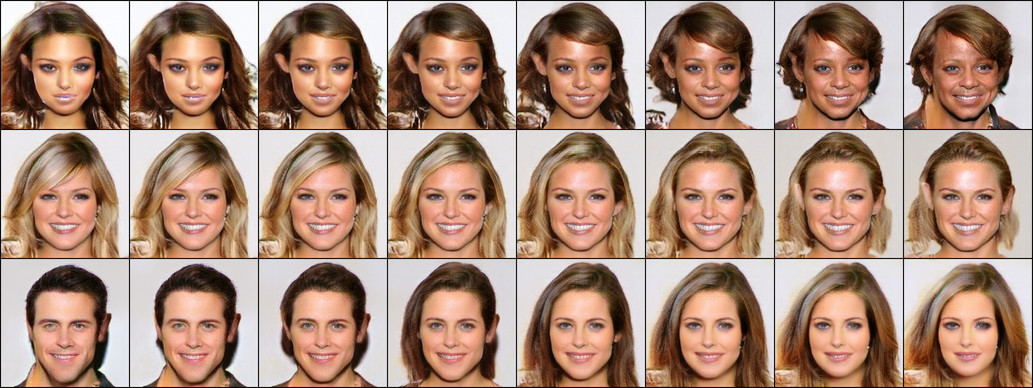}
            \caption{Attribute (age, bangs, gender) interpolation in feature space.}
        \end{subfigure}\vspace{8pt}
        \begin{subfigure}{\linewidth} 
            \centering
            \includegraphics[width=\linewidth]{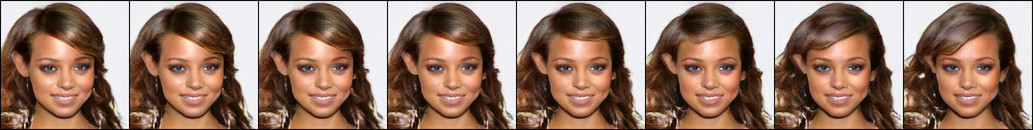}
            \caption{Attribute (age) interpolation in latent space.}
        \end{subfigure}
    \end{subfigure}
    \end{center}
    \caption{Pose and attribute interpolation.}
    \label{fig:interp_faces}
\end{figure}

\subsection{Single Source and Intra-Class Diversity} \label{subsec:diversity}
Our method is able to generate a range of samples from a single source image by randomly sampling the latent vector. This single source diversity is demonstrated in Figure \ref{fig:vis_dist}. The generated samples match the semantic features of the source image, but varying the latent vector results in structural changes, such as the pose and number of flowers present. Intra-class diversity is demonstrated in Figure \ref{fig:fixed_z} by fixing the latent vector and sampling various features from the same class. Due to the fixed latent vector, the structural information is consistent, while the sampling of different features results in fine-grained intra-class differences, such as colour. Again, all source images are from novel classes.

\begin{figure}[!t]
    \begin{center}
    \begin{subfigure}{0.74\linewidth}
    \centering
        \begin{subfigure}{.08333333\linewidth} 
            \centering
            \includegraphics[width=\linewidth]{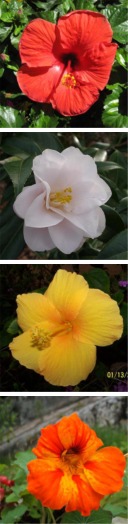}
            \caption*{Source.}
        \end{subfigure}\hspace{4pt}
        \begin{subfigure}{.416666666\linewidth} 
            \centering
            \includegraphics[width=\linewidth]{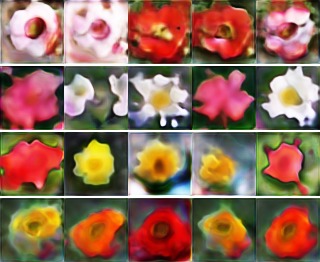}
            \caption*{DAGAN \cite{antoniou2017data}.}
        \end{subfigure}\hspace{4pt}
        \begin{subfigure}{.4166666666\linewidth} 
            \centering
            \includegraphics[width=\linewidth]{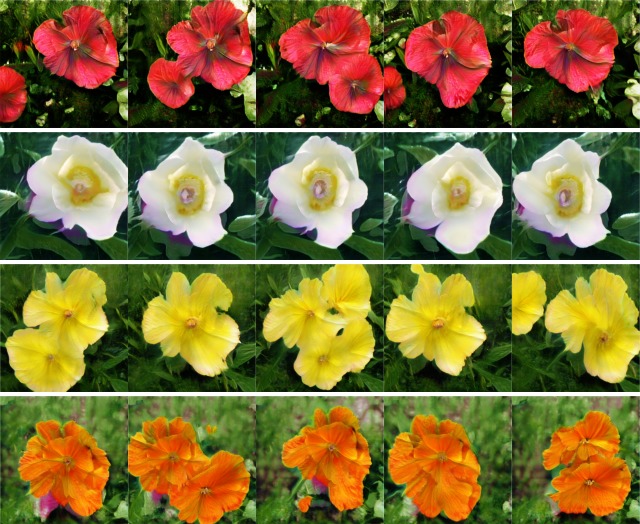}
            \caption*{Ours: OpenGAN.}
        \end{subfigure}
    \end{subfigure}
    \end{center}
    \caption{Fixed source image per row with random latent vectors. DAGAN samples show significant semantic variation, while OpenGAN samples do not. }
    \label{fig:split}
\end{figure}

\subsection{Latent and Feature Space Interpolation} \label{subsec:interpolation}

A two-dimensional interpolation between two latent vectors (horizontal direction) and two feature embeddings (vertical direction) is shown in Figure \ref{fig:flowers_2d}. 
The generated samples are required to contain the semantic information encoded in the given feature embedding. As such, interpolation in latent space with a fixed feature results in plausible transformations in the image space, without changes in the fine-grained semantic content. This is unlike latent space interpolation in conventional cGANs, which by design results in intra-class semantic variations.

By training a classifier to predict the binary pose and attribute labels of CelebA, we are able to compute pose/attribute mean latent and feature vectors. If a given attribute is encoded, traversing the line that connects the mean positive and negative vectors will vary that attribute in the image space. As seen in Figure \ref{fig:interp_faces}, pose information is encoded only in the latent space, with no pose change seen when interpolating between the mean feature vectors. Conversely, attributes such as age, gender and hair style are encoded only in feature space.

\begin{figure}[!t]
    \begin{minipage}{.36\textwidth}
        \begin{center}
        \begin{subfigure}{0.48\linewidth} 
            \includegraphics[width=\linewidth]{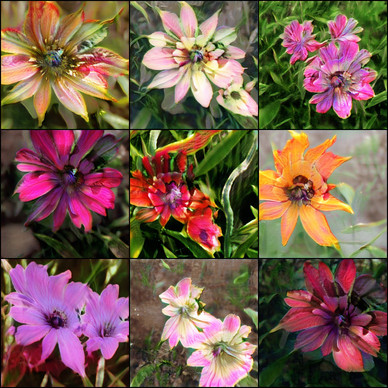}
        \end{subfigure}\hspace{4pt}%
        \begin{subfigure}{0.48\linewidth} 
            \includegraphics[width=\linewidth]{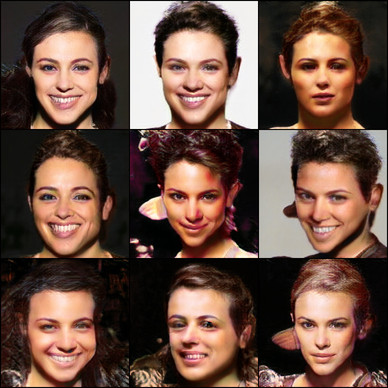}
        \end{subfigure}
        \end{center}
        \caption{Random sampling of the feature space.}
        \label{fig:random_sample}
    \end{minipage}\hspace{14pt}%
    \begin{minipage}{.56\textwidth}
        \begin{center}
        \begin{subfigure}{\linewidth} 
        \centering
            \begin{subfigure}{0.46\linewidth} 
                \centering
                \includegraphics[width=\linewidth]{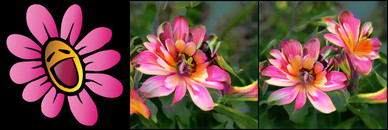}
            \end{subfigure}\hspace{2pt}%
            \begin{subfigure}{0.46\linewidth} 
                \centering
                \includegraphics[width=\linewidth]{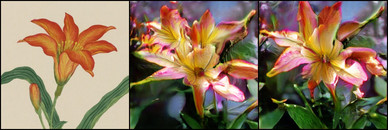}
            \end{subfigure}\vspace{1pt}
            \begin{subfigure}{0.46\linewidth} 
                \centering
                \includegraphics[width=\linewidth]{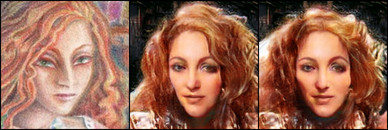}
            \end{subfigure}\hspace{2pt}%
            \begin{subfigure}{0.46\linewidth} 
                \centering
                \includegraphics[width=\linewidth]{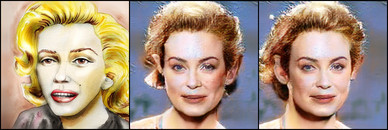}
            \end{subfigure}
        \end{subfigure}
        \end{center}
        \caption{Generating from out-of-domain source images.}
        \label{fig:art}
    \end{minipage}
\end{figure}

\subsection{Split of Information in Latent and Feature Spaces} \label{subsec:split}
As seen in Figures \ref{fig:latentvsfeature} and \ref{fig:interp_faces}, all discriminative semantic information is encoded in the feature space and only non-discriminative structural information is encoded in the latent space.
In Figure \ref{fig:split}, we show that this clean split does not exist in encoder-decoder style image-conditional GANs, such as DAGAN \cite{antoniou2017data}. We train a DAGAN model using the official implementation on Flowers102. It can be seen that for a fixed source image, DAGAN samples undesirably show significant semantic variation (e.g. the colour of the flower) when varying only the latent space, while OpenGAN samples show no discriminative semantic variation.

\subsection{Random Feature Space Sampling} \label{subsec:random_sample}
Conventional GANs are able to generate data by randomly sampling the latent space without any external inputs. Our generator is trained not only with latent space sampling, but also feature space sampling. Figure \ref{fig:random_sample} shows that new data can be generated by randomly selecting both the latent and feature vectors. The generated samples are diverse, as well as visually and semantically plausible.

\subsection{Out-of-Domain Source Images} \label{subsec:domain}

We investigate the use of out-of-domain source images, such as paintings and digital art, that have similar semantic content as the training images. As seen in Figure \ref{fig:art}, the fake samples match the semantic features of the source images. This shows that the metric learning model is able to extract relevant information, despite the domain shift.

\subsection{OpenGAN for Data Augmentation} \label{subsec:data_aug}
In this section, we demonstrate the usefulness of samples generated by OpenGAN to the downstream application of data augmentation for classification. As a baseline, we train a Resnet18 \cite{7780459} classifier on 500 novel (i.e. outside of the OpenGAN training distribution) CelebA classes, using 1, 2, 5 and 10 training examples per class. The same test set is used for all experiments.  
To train the classifier with data augmentation, we first sample a batch of real images from the training data set and perform an optimisation step on the classifier. Using the metric features extracted from the sampled real images, a batch of fake images is generated, which is used to perform another optimisation step on the classifier. 
The randomised generation of fake images and classifier optimisation step is repeated using the same batch of real features until the desired ratio of fake-to-real data is achieved. A new batch of real images is then sampled and the process repeats.
We find that adding small random perturbations to the real features before generating fake data can be beneficial. The perturbations are Gaussian noise with a zero mean and standard deviation of $\eta \, \sigma_F$, where $\eta$ is a scaling term and $\sigma_F$ is the standard deviation of the real features across all examples and dimensions.

For each number of real samples per class, we experiment with fake-to-real data ratios of 1 through 5 and $\eta$ values of 0, 1.5 and 2. The best performing experiments for each number of real samples per class are shown in Table \ref{table:data_aug}. Data augmentation results in a significant improvement in classification performance, despite the classes being from outside of the OpenGAN training distribution.

\begin{table}[t]
\setlength{\tabcolsep}{8pt}
\centering
\caption{CelebA data augmentation using OpenGAN samples.}
\label{table:data_aug}
\begin{tabular}{lccccc}
\toprule 
& Real Per Class & Fake Per Real & $\eta$ & Test Acc. (\%) \\
\cmidrule(lr){2-5}
Baseline & 1 & 0 & - & 2.71 \\
With Data Aug. & 1 & 5 & 2 & \textbf{12.13} \\
\midrule
Baseline & 2 & 0 & - & 7.47 \\
With Data Aug. & 2 & 4 & 1.5 & \textbf{22.70} \\
\midrule
Baseline & 5 & 0 & - & 25.98 \\
With Data Aug. & 5 & 3 & 1.5 & \textbf{51.81} \\
\midrule
Baseline & 10 & 0 & - & 52.69 \\
With Data Aug. & 10 & 2 & 1.5 & \textbf{71.98} \\
\bottomrule
\end{tabular}
\end{table}

\section{Conclusion}

In this paper, we proposed a generative adversarial network that is conditioned on per-sample feature embeddings drawn from a metric space. Such an approach allows the generation of samples that are semantically similar to a given source image. Our method is able to generate data from novel classes that are outside of the training distribution. 
We demonstrated that interpolation in the feature and latent spaces results in semantically plausible samples, with the feature space encoding fine-grained semantic information and the latent space encoding structural information. Finally, generated samples can be used to significantly improve classification performance through data augmentation.

\section{Acknowledgements}
This research was supported by the Australian Research Council Centre of Excellence for Robotic Vision (project number CE140100016).

%
%
\bibliographystyle{splncs04}

\clearpage
\appendix

\section{Additional Implementation Details}
\subsection{Network Architecture}
\begin{figure}[t]
    \begin{center}
    \begin{subfigure}[b]{0.42\linewidth} 
        \includegraphics[width=\linewidth]{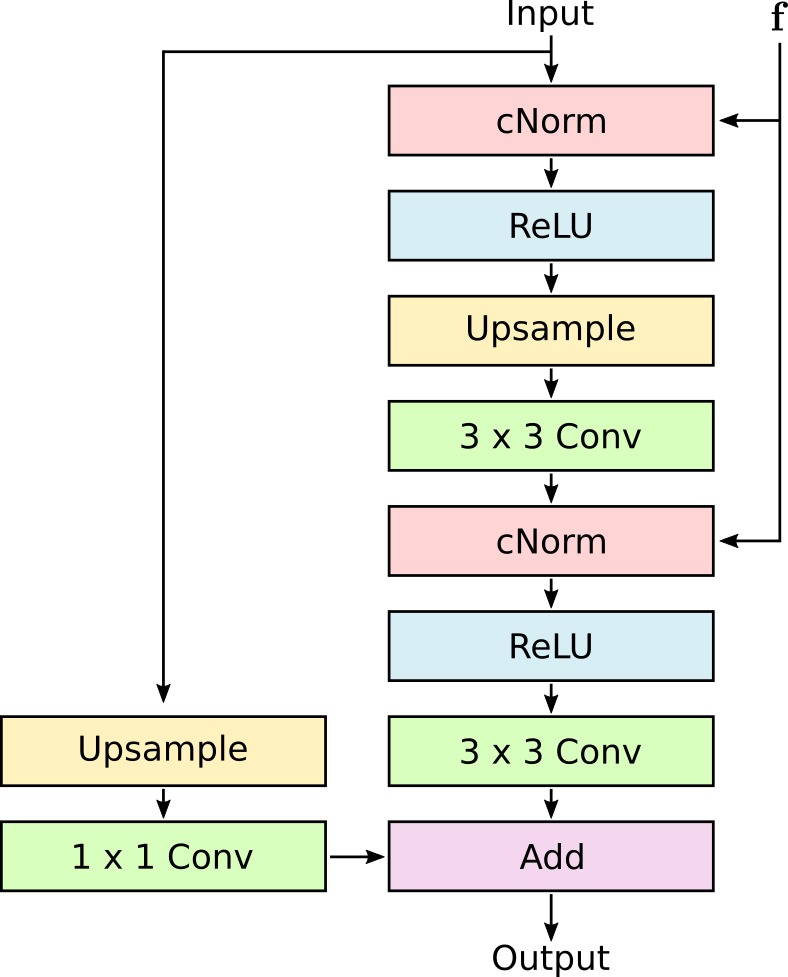}
        \caption{ResBlock Up}
        \label{subfig:resup}
    \end{subfigure}\hspace{40pt}%
    \begin{subfigure}[b]{0.42\linewidth} 
        \includegraphics[width=\linewidth]{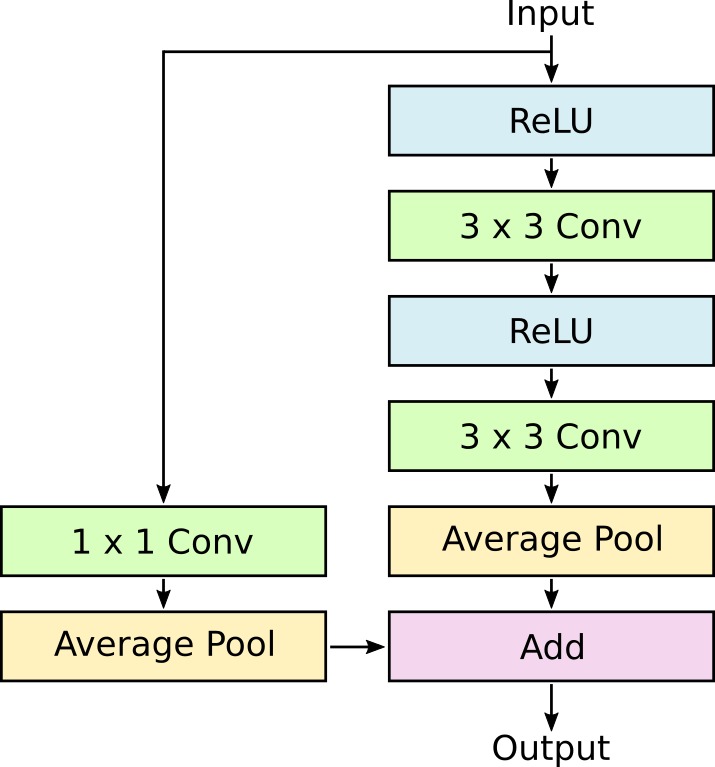}
        \caption{ResBlock Down}
        \label{subfig:resdown}
    \end{subfigure}
    \end{center}
    \caption{Structure of residual blocks. Upsampling layers (nearest neighbour) and downsampling layers (average pooling) change the scale by a factor of two.}
    \label{fig:res_blocks}
\end{figure}

The generator and discriminator architectures are shown in Tables \ref{subtable:gen} and \ref{subtable:dis}, respectively. Spectral normalisation \cite{miyato2018spectral} is used on all weights, except in the feature embedding conditional normalisation (\textit{cNorm}) blocks. The structure of the up-sampling and down-sampling residual blocks (\textit{ResBlocks}) are shown in Figures \ref{subfig:resup} and \ref{subfig:resdown}, respectively. The baseline models follow the same architecture, with the only difference being the calculation of the normalisation layer scale and bias terms. The non-conditional baseline has no normalisation layers, while the class-conditional baseline uses a single conditioning embedding per class.
A standard Resnet18 network \cite{7780459} is used for the feature extractor, with the softmax layer and class-specific fully connected layer removed.

\begin{table}[H]
\begin{center}
\begin{subtable}[b]{0.45\linewidth}
\centering
\begin{tabular}{c}
\toprule 
$\mathbf{z} \in \mathbb{R}^{128} \sim \mathcal{N}(0,1)$ \\
$\mathbf{f} \in \mathbb{R}^{512}, \mathbf{f} = F(\mathbf{x}), \mathbf{x} \sim p_d$ \\
\midrule
Linear $128 \to 512 \times 4 \times 4$  \\
\midrule
ResBlock Up $512 \to 512$ \\ 
\midrule
ResBlock Up $512 \to 256$ \\ 
\midrule
ResBlock Up $256 \to 256$ \\ 
\midrule
ResBlock Up $256 \to 128$ \\
\midrule
Self-Attention Block \\
\midrule
ResBlock Up $128 \to 64$ \\ 
\midrule
ResBlock Up $64 \to 32$ \\ 
\midrule
Normalisation, ReLU \\
\midrule
$3 \times 3$ Conv $32 \to 3$\\
\midrule
Tanh\\
\bottomrule
\end{tabular}
\caption{Generator.}
\label{subtable:gen}
\end{subtable}\hspace{15pt}%
\begin{subtable}[b]{0.45\linewidth}
\centering
\begin{tabular}{c}
\toprule 
$\mathbf{x} \in \mathbb{R}^{256 \times 256 \times 3}$\\
\midrule
$3 \times 3$ Conv $3 \to 32$\\
\midrule
ResBlock Down $32 \to 64$ \\ 
\midrule
ResBlock Down $64 \to 128$ \\ 
\midrule
Self-Attention Block \\
\midrule
ResBlock Down $128 \to 256$ \\ 
\midrule
ResBlock Down $256 \to 256$ \\
\midrule
ResBlock Down $256 \to 512$ \\ 
\midrule
ResBlock Down $512 \to 512$ \\ 
\midrule
cNorm, ReLU \\
\midrule
$4 \times 4$ Conv $512 \to 1$\\
\bottomrule
\end{tabular}
\caption{Discriminator.}
\label{subtable:dis}
\end{subtable}
\end{center}
\caption{Network architectures to generate $256 \times 256$ samples. The real data distribution is denoted as $p_d$.}
\end{table}

\subsection{Attribute Interpolation} \label{subsec:attributes}
In this section, we describe the method used to perform attribute and pose interpolation in more detail. The binary attribute labels are taken directly from the Celeba dataset \cite{liu2015faceattributes}. Pose labels, for example left facing, right facing and forward facing, are found by using the facial landmark locations of the data. A Resnet18 network \cite{7780459} is trained as a multi-label classifier on the training data using binary cross-entropy loss. To perform the interpolation, the positive and negative mean latent vectors and feature embeddings are found for each attribute and pose. This is achieved by predicting the attribute and pose labels of generated samples and grouping the associated latent and feature vectors. The unit vector that points from the positive group to the negative group for all attributes and poses are found for both the latent and feature spaces. To perform interpolation, an image is first generated using a source image and randomly sampled latent vector. For interpolation of a given attribute in the feature space, for example, the scaled attribute unit vector is added to the starting feature embedding. Samples are generated across a range of unit vector scaling terms.

\section{Additional Results}

In this section, we include additional results to further evaluate the performance of our proposed method.

\subsubsection{One-Shot Image Generation}
\begin{figure}[!t]
    \begin{center}
    \begin{subfigure}{\linewidth} 
        \begin{subfigure}{\linewidth} 
            \includegraphics[width=\linewidth]{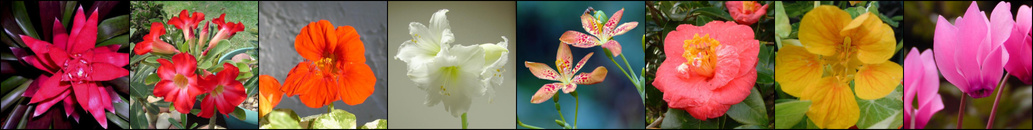}
        \end{subfigure}\vspace{1pt}
        \begin{subfigure}{\linewidth} 
            \includegraphics[width=\linewidth]{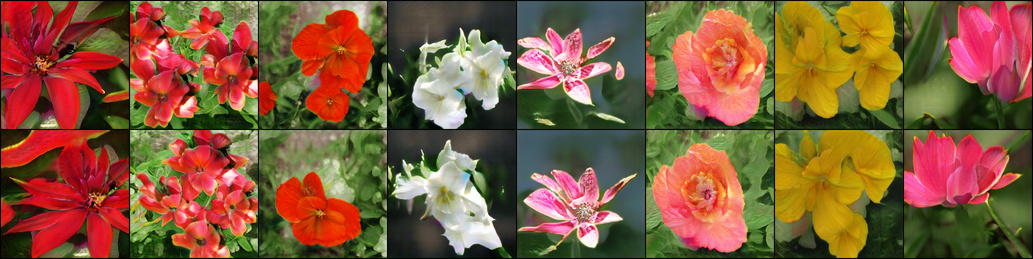}
        \end{subfigure}\vspace{5pt}
        \begin{subfigure}{\linewidth} 
            \includegraphics[width=\linewidth]{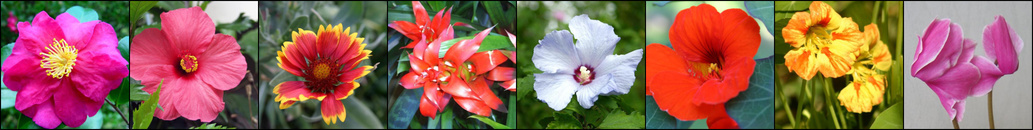}
        \end{subfigure}\vspace{1pt}
        \begin{subfigure}{\linewidth} 
            \includegraphics[width=\linewidth]{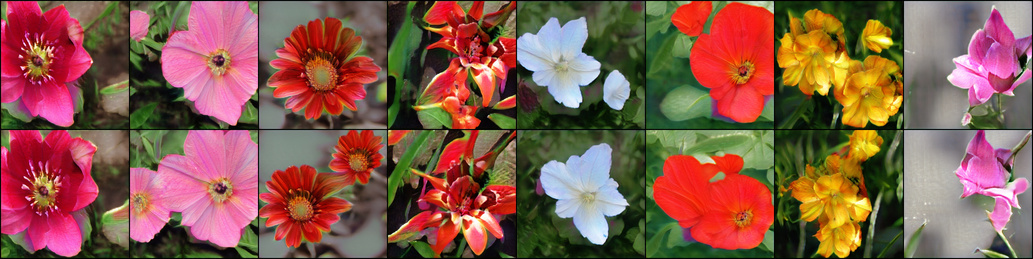}
        \end{subfigure}
    \end{subfigure}
    \end{center}
    \caption{Flowers102 novel class real source images (top row of each section) and resultant generated images (bottom two rows of each section). Although the species are not present during training, the fake images match the features of the real source images.}
    \label{fig:flowers_vis_dist}
\end{figure}

\begin{figure}[!t]
    \begin{center}
    \begin{subfigure}{\linewidth} 
        \begin{subfigure}{\linewidth} 
            \includegraphics[width=\linewidth]{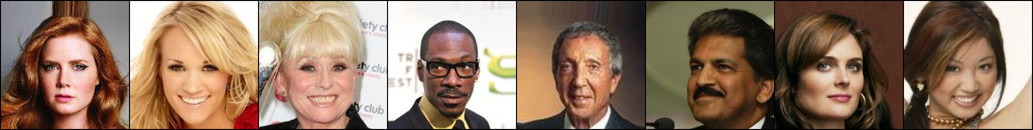}
        \end{subfigure}\vspace{1pt}
        \begin{subfigure}{\linewidth} 
            \includegraphics[width=\linewidth]{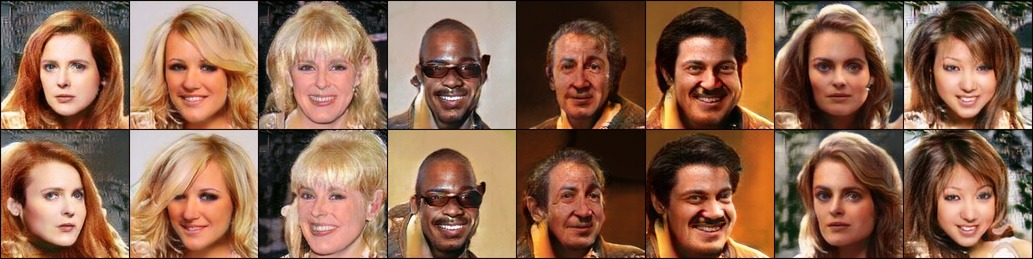}
        \end{subfigure}\vspace{5pt}
        \begin{subfigure}{\linewidth} 
            \includegraphics[width=\linewidth]{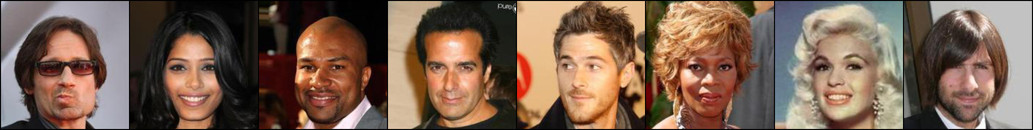}
        \end{subfigure}\vspace{1pt}
        \begin{subfigure}{\linewidth} 
            \includegraphics[width=\linewidth]{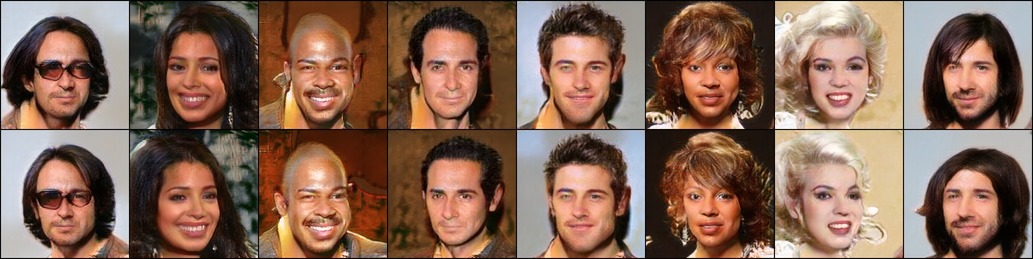}
        \end{subfigure}
    \end{subfigure}
    \end{center}
    \caption{Celeba novel class real source images (top row of each section) and resultant generated images (bottom two rows of each section). Although the identities are not present during training, the fake images match the features of the real source images.}
    \label{fig:faces_vis_dist_supp}
\end{figure}

Figures \ref{fig:flowers_vis_dist} (Flowers102 \cite{Nilsback08}) and \ref{fig:faces_vis_dist_supp} (Celeba \cite{liu2015faceattributes}) show samples generated when the generator is conditioned on feature embeddings extracted from novel class source images. Despite the classes being from outside of the training distribution, the generated samples match the semantic features found in the source images.

\subsubsection{Attribute Interpolation}
\begin{figure}[!t]
    \begin{center}
    \begin{subfigure}{\linewidth}
    \begin{subfigure}{\linewidth} 
        \begin{subfigure}{.058015\linewidth} 
            \includegraphics[width=\linewidth]{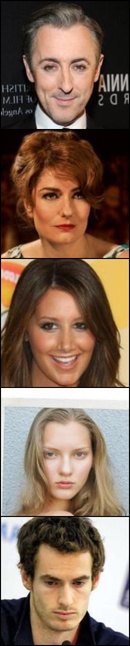}
        \caption*{Real.}
        \end{subfigure}\hspace{3pt}
        \begin{subfigure}{0.460993\linewidth} 
            \includegraphics[width=\linewidth]{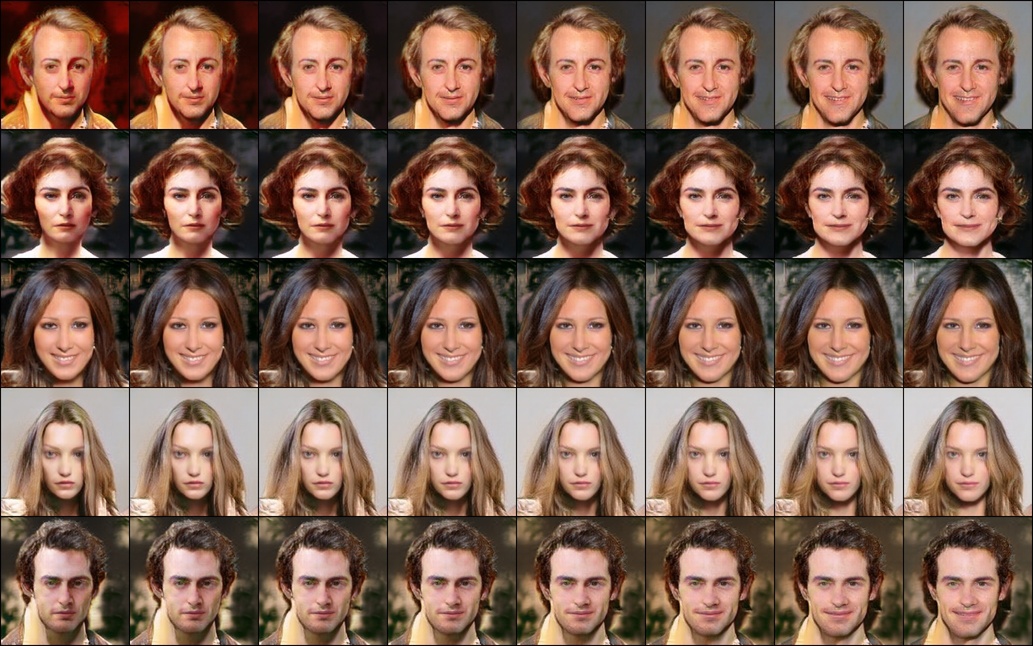}
        \caption*{Feature space interpolation: pose.}
        \end{subfigure}\hspace{3pt}
        \begin{subfigure}{0.460993\linewidth} 
            \includegraphics[width=\linewidth]{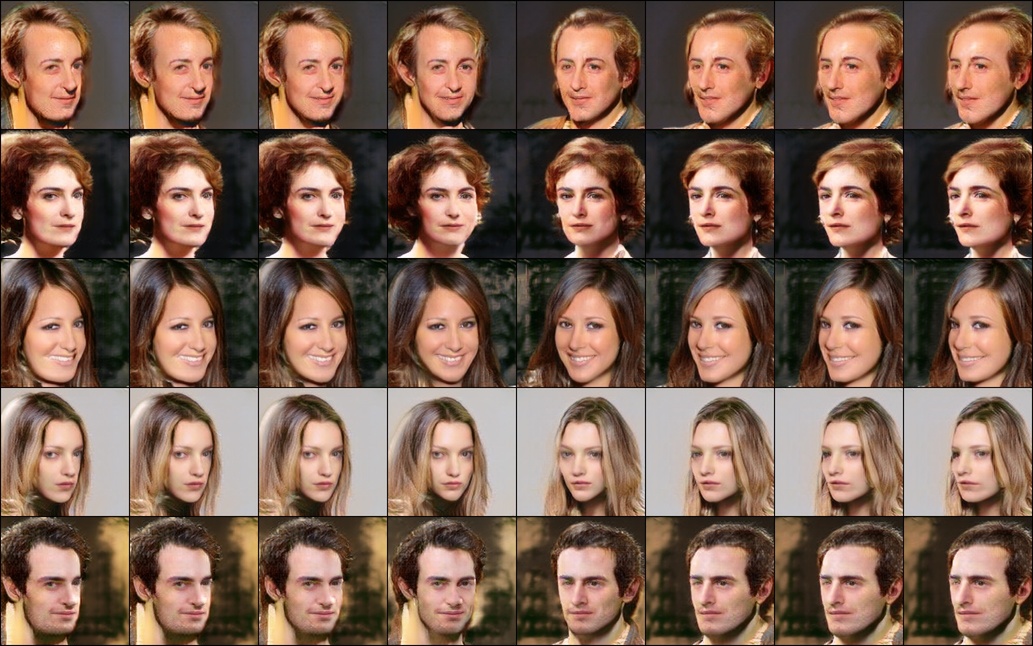}
        \caption*{Latent space interpolation: pose.}
        \end{subfigure}
    \end{subfigure}\vspace{6pt}\\
    \begin{subfigure}{\linewidth} 
        \begin{subfigure}{.058015\linewidth} 
            \includegraphics[width=\linewidth]{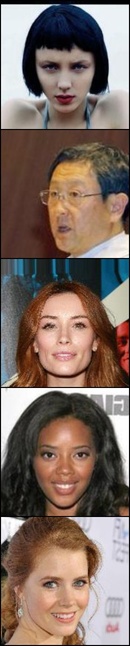}
        \caption*{Real.}
        \end{subfigure}\hspace{3pt}
        \begin{subfigure}{0.460993\linewidth} 
            \includegraphics[width=\linewidth]{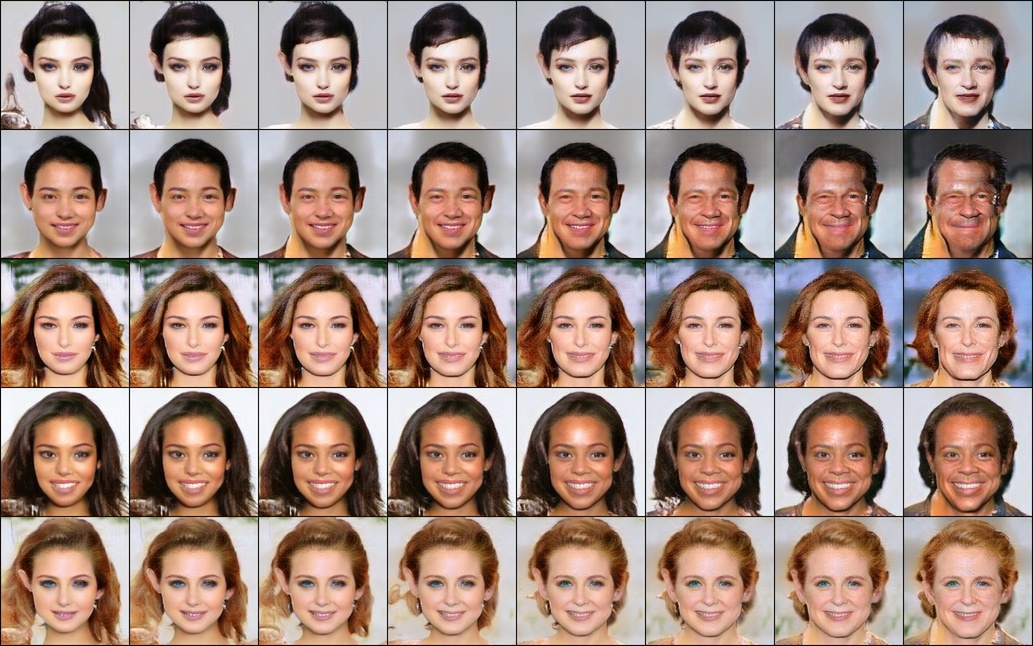}
        \caption*{Feature space interpolation: age.}
        \end{subfigure}\hspace{3pt}
        \begin{subfigure}{0.460993\linewidth} 
            \includegraphics[width=\linewidth]{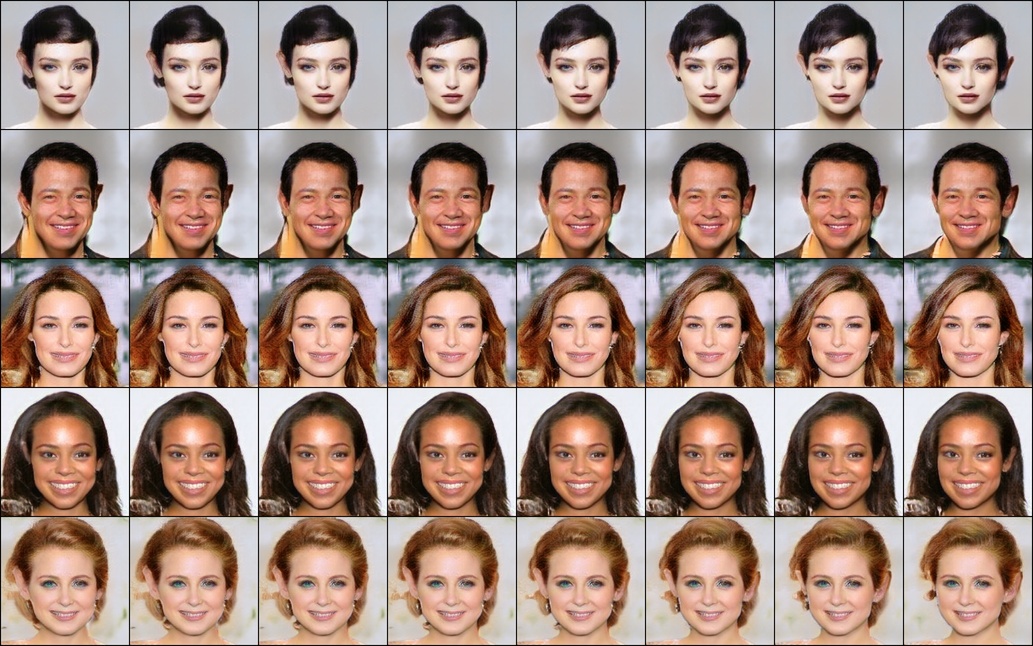}
        \caption*{Latent space interpolation: age.}
        \end{subfigure}
    \end{subfigure}\vspace{6pt}\\
    \begin{subfigure}{\linewidth} 
        \begin{subfigure}{.058015\linewidth} 
            \includegraphics[width=\linewidth]{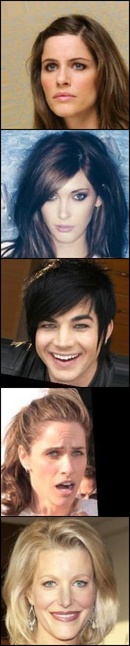}
        \caption*{Real.}
        \end{subfigure}\hspace{3pt}
        \begin{subfigure}{0.460993\linewidth} 
            \includegraphics[width=\linewidth]{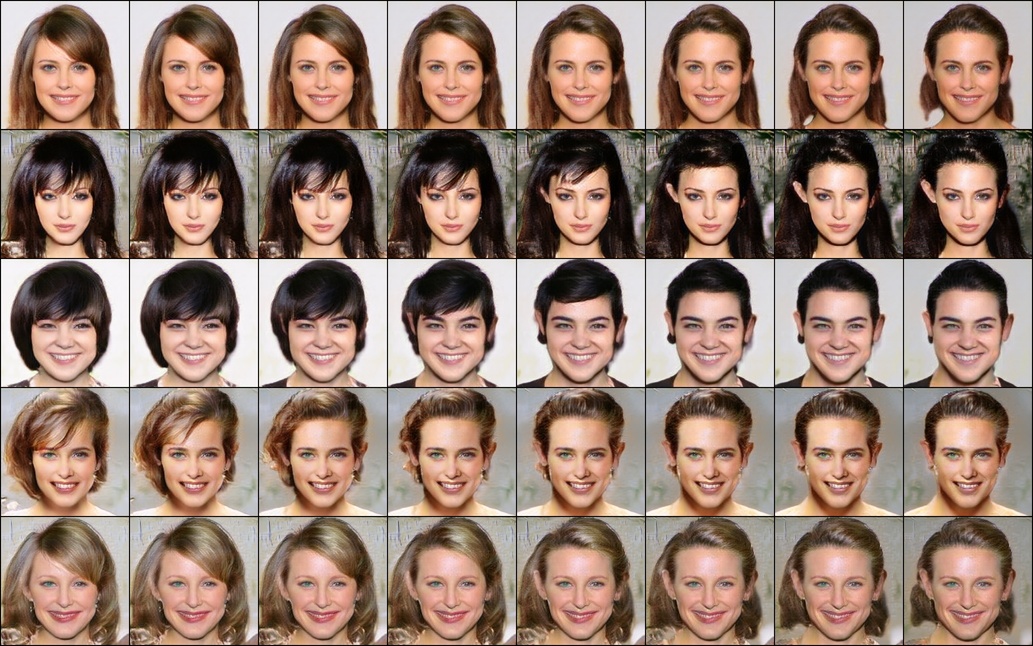}
        \caption*{Feature space interpolation: bangs.}
        \end{subfigure}\hspace{3pt}
        \begin{subfigure}{0.460993\linewidth} 
            \includegraphics[width=\linewidth]{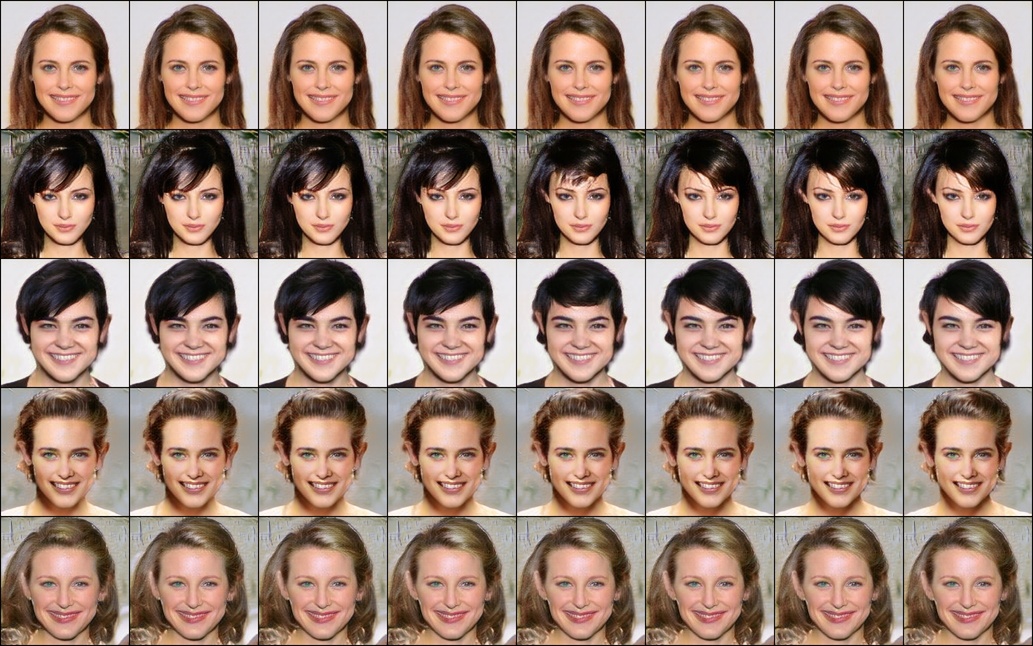}
        \caption*{Latent space interpolation: bangs.}
        \end{subfigure}
    \end{subfigure}\vspace{6pt}\\
    \begin{subfigure}{\linewidth} 
        \begin{subfigure}{.058015\linewidth} 
            \includegraphics[width=\linewidth]{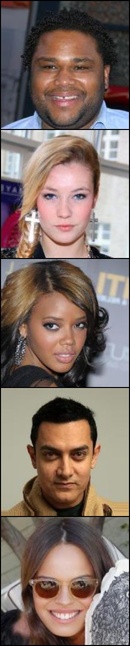}
        \caption*{Real.}
        \end{subfigure}\hspace{3pt}
        \begin{subfigure}{0.460993\linewidth} 
            \includegraphics[width=\linewidth]{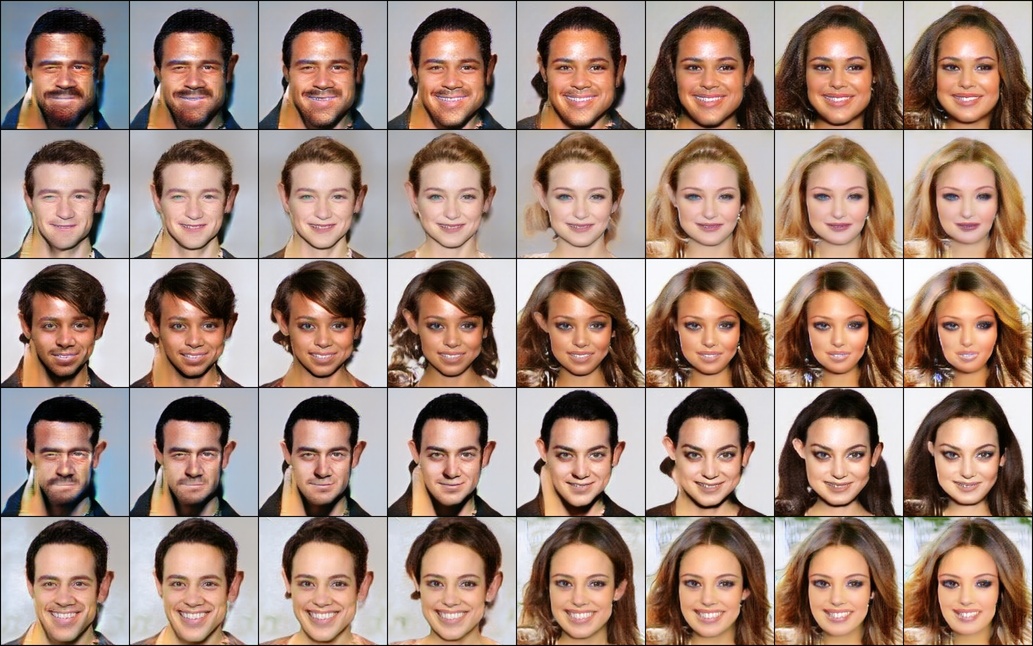}
        \caption*{Feature space interpolation: gender.}
        \end{subfigure}\hspace{3pt}
        \begin{subfigure}{0.460993\linewidth} 
            \includegraphics[width=\linewidth]{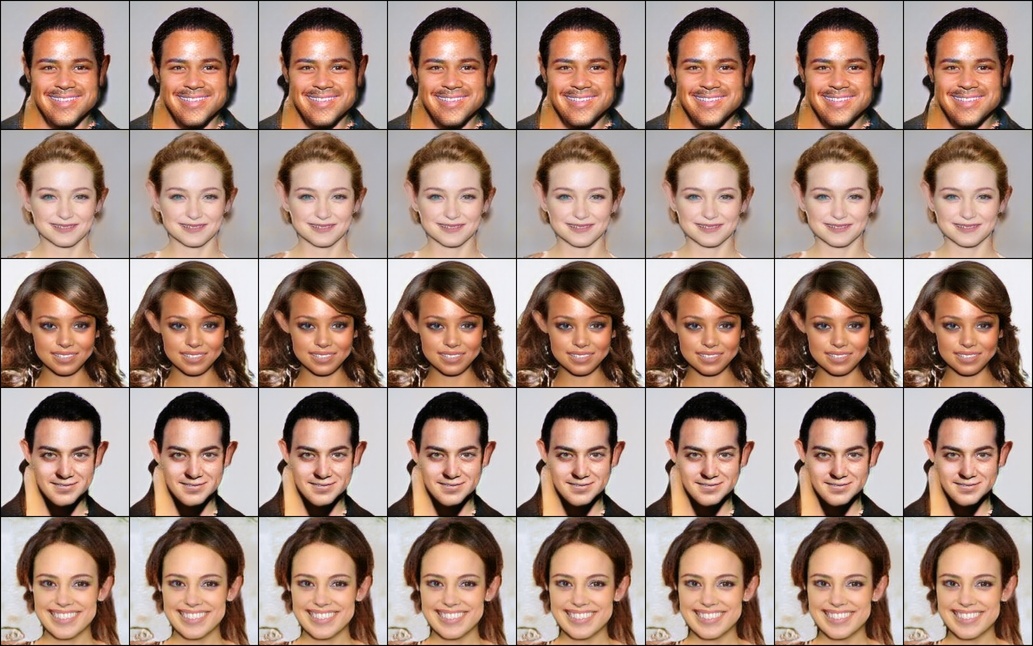}
        \caption*{Latent space interpolation: gender.}
        \end{subfigure}
    \end{subfigure}
    \end{subfigure}
    \end{center}
    \caption{Pose is encoded only in the latent space, while age, hairstyle (bangs) and gender are encoded only in the feature space.}
    \label{fig:attr_interp}
\end{figure}

The method detailed in Section \ref{subsec:attributes} is used to perform interpolation between poses and attributes in Figure \ref{fig:attr_interp}. Interpolations are shown in both the feature space and latent space. It can be seen that pose interpolation in the feature space has no impact on the pose in the generated samples. This indicates that pose is encoded only in the latent space. By contrast, attributes (age, bangs and gender) are encoded only in the feature space. Further pose and attribute interpolation can be observed in the included videos.

\subsubsection{Random Interpolation}
\begin{figure}[!t]
    \begin{center}
    \begin{subfigure}{\linewidth} 
        \includegraphics[width=\linewidth]{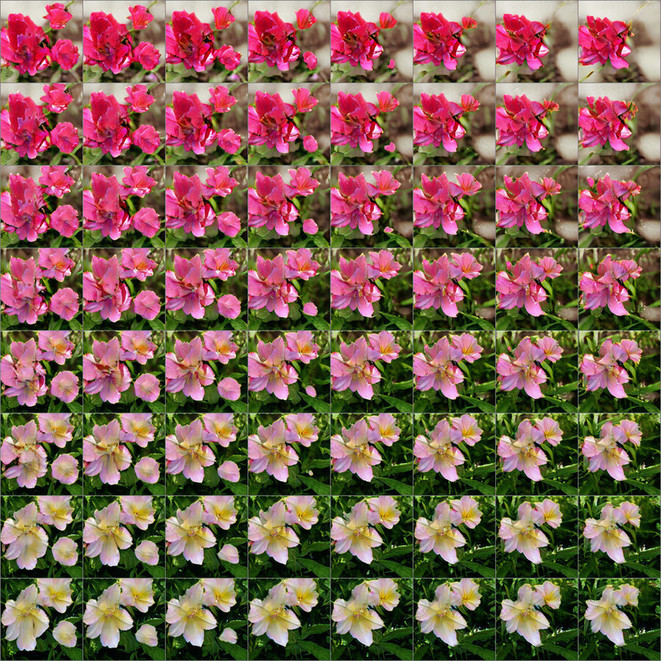}
    \end{subfigure}
    \end{center}
    \caption{Interpolation between two latent vectors (horizontal) and two feature embeddings (vertical). Feature embeddings are from novel classes.}
    \label{fig:2d_interp2}
\end{figure}

\begin{figure}[!t]
    \begin{center}
    \begin{subfigure}{\linewidth} 
        \includegraphics[width=\linewidth]{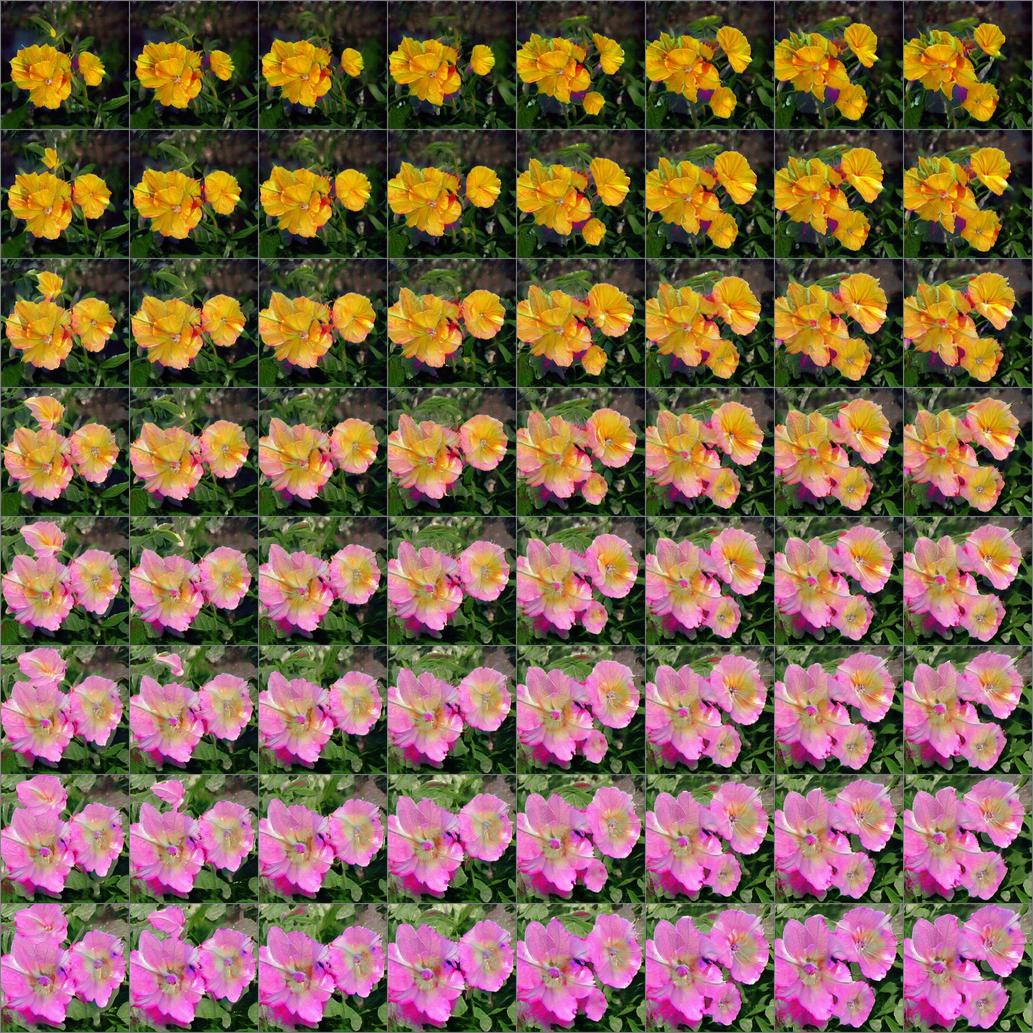}
    \end{subfigure}
    \end{center}
    \caption{Interpolation between two latent vectors (horizontal) and two feature embeddings (vertical). Feature embeddings are from novel classes.}
    \label{fig:2d_interp1}
\end{figure}

Figures \ref{fig:2d_interp2} and \ref{fig:2d_interp1} show interpolation between two random latent vectors (horizontal direction) and two sampled novel class feature embeddings (vertical direction). It can be seen that only structural information changes when the latent vector is varied, while semantic information changes when the feature embedding is varied. Further random interpolation can be observed in the included videos.

\subsubsection{Random Feature Sampling}
\begin{figure}[!t]
    \begin{center}
    \begin{subfigure}{0.95\linewidth} 
        \includegraphics[width=\linewidth]{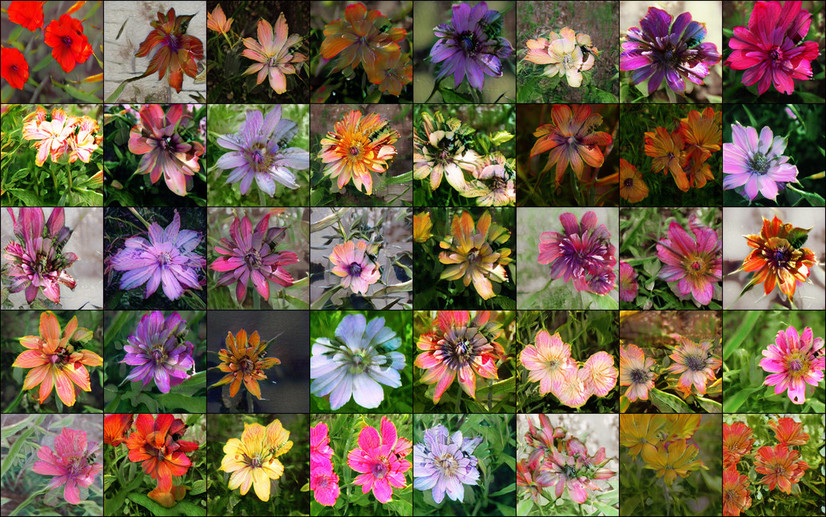}
    \end{subfigure}\vspace{1pt}\\
    \begin{subfigure}{0.95\linewidth} 
        \includegraphics[width=\linewidth]{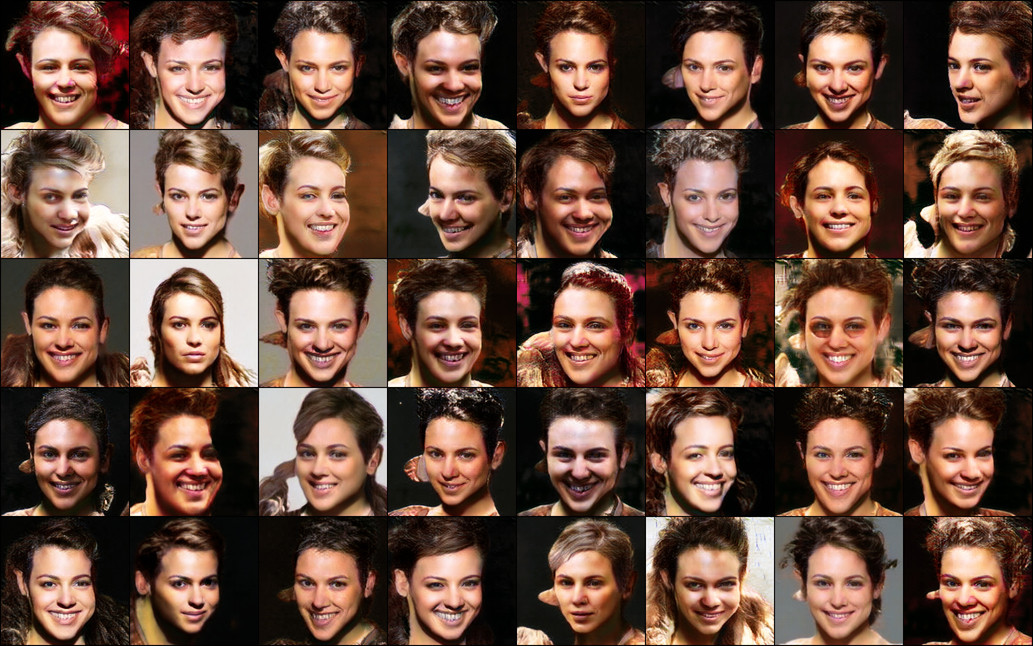}
    \end{subfigure}
    \end{center}
    \caption{Uncurated images generated by randomly sampling the metric feature space and latent space. No class-level information or other labels are used to generate these samples.}
    \label{fig:random_sampling}
\end{figure}

Further samples generated by randomly sampling the metric feature space are shown in Figure \ref{fig:random_sampling}. Features are sampled using a single mean and standard deviation across all embedding dimensions. No class-level information or other labels are utilised. 

\end{document}